\documentclass[review]{elsarticle}
\usepackage{subfig}
\usepackage{array}
\usepackage{color} % change colors
\usepackage{bbding} % check
\usepackage{multirow}
\usepackage{amssymb}
\usepackage{amsmath}
\usepackage{hyperref}

\journal{Pattern Recognition}

\begin{document}
\begin{frontmatter}

%% Title, authors and addresses

%% use the tnoteref command within \title for footnotes;
%% use the tnotetext command for theassociated footnote;
%% use the fnref command within \author or \address for footnotes;
%% use the fntext command for theassociated footnote;
%% use the corref command within \author for corresponding author footnotes;
%% use the cortext command for theassociated footnote;
%% use the ead command for the email address,
%% and the form \ead[url] for the home page:
%% \title{Title\tnoteref{label1}}
%% \tnotetext[label1]{}
%% \author{Name\corref{cor1}\fnref{label2}}
%% \ead{email address}
%% \ead[url]{home page}
%% \fntext[label2]{}
%% \cortext[cor1]{}
%% \affiliation{organization={},
%%             addressline={},
%%             city={},
%%             postcode={},
%%             state={},
%%             country={}}
%% \fntext[label3]{}

\title{Single image super-resolution based on trainable feature matching attention network}

\author[qz]
{Qizhou Chen}
\ead{chen_qizhou@outlook.com}

\author[sq]
{Qing Shao\corref{cor1}}
\ead{qshao@usst.edu.cn} 
\cortext[cor1]{Corresponding author}

\affiliation[qz,sq]
{
    organization={School of Optoelectronic Information and Computer Engineering},
    addressline={University of Shanghai for Science and Technology},
    city={Shanghai},
    postcode={200093},
    % state={China},
    country={China}
}

\begin{abstract}

    Convolutional Neural Networks (CNNs) have been widely employed for 
    image Super-Resolution (SR) in recent years. 
    Various techniques enhance SR performance by altering CNN structures 
    or incorporating improved self-attention mechanisms.
    Interestingly, these advancements share a common trait.
    Instead of explicitly learning high-frequency details, they 
    learn an implicit feature processing mode that utilizes 
    weighted sums of a feature map's own elements for reconstruction, 
    akin to convolution and non-local.
    In contrast, early dictionary-based approaches learn feature 
    decompositions explicitly to match and rebuild Low-Resolution (LR) features.
    Building on this analysis, we introduce Trainable Feature Matching (TFM) 
    to amalgamate this explicit feature learning into CNNs, augmenting their 
    representation capabilities.
    Within TFM, trainable feature sets are integrated to explicitly learn 
    features from training images through feature matching. 
    Furthermore, we integrate non-local and channel attention into 
    our proposed Trainable Feature Matching Attention Network (TFMAN) 
    to further enhance SR performance.
    To alleviate the computational demands of non-local operations, 
    we propose a streamlined variant called Same-size-divided 
    Region-level Non-Local (SRNL). SRNL conducts non-local computations 
    in parallel on blocks uniformly divided from the input feature map.
    The efficacy of TFM and SRNL is validated through ablation studies 
    and module explorations. We employ a recurrent convolutional network 
    as the backbone of our TFMAN to optimize parameter utilization. 
    Comprehensive experiments on benchmark datasets demonstrate that 
    TFMAN achieves superior results in most comparisons while using 
    fewer parameters. 
    The code is available at \url{https://github.com/qizhou000/tfman}.

    \end{abstract}

\begin{keyword}
Super-resolution, feature matching, non-local, 
recurrent convolutional neural network, 
deep learning
\end{keyword}

\end{frontmatter}

\section{Introduction}
% The goal of single image super-resolution (SISR) is to 
% recover a high-resolution (HR) image from a someway down-sampled 
% low-resolution (LR) counterpart.
% Obviously, the problem is ill-posed since it attempts to build a mapping from a 
% low-dimensional manifold to a high-dimensional one.
% SR not only has practical significance in medical imaging, security and 
% surveillance, but improves the performance of many advanced visual tasks
% like object detection and semantic segmentation.
% It is an open-ended question and remains challenging.

Single Image Super-Resolution (SISR) aims to restore a High-Resolution (HR) 
image from a down-sampled Low-Resolution (LR) version. 
This task is inherently challenging as it involves mapping from 
a lower-dimensional manifold to a higher-dimensional one, 
presenting an ill-posed problem. Despite its ill-posed nature, 
Super-Resolution (SR) holds not only practical importance in fields 
such as medical imaging, security, and surveillance but also 
enhances the outcomes of various advanced visual applications 
like object detection and semantic segmentation. 
It is an open-ended question and remains challenging.

Since the introduction of SRCNN \cite{SR:SRCNN}, CNN-based 
methods have greatly improved SR performance. 
Most methods aim to ensure consistency with 
ground truth, seen in approaches like VDSR \cite{SR:VDSR} 
and RCAN \cite{SR:RCAN}, focusing on improving SR metrics 
(PSNR and SSIM). Some, like SRFeat \cite{SR:SRFeat}, 
prioritize visual authenticity using perceptual and 
adversarial loss against over-smoothing. 
Patch loss \cite{J_PR_SR_PatchLoss}, similarly, 
enhances restoration by minimizing multiscale 
similarity of image patches. Of course, these losses may be 
harmful for models to get higher PSNR and SSIM.
Moreover, due to the ill-posed nature of SR tasks, 
diversified reconstruction is also a noteworthy point, 
such as standard flow-based SRFlow \cite{SR:SRFlow}.
In this paper, we specifically concentrate on the first aspect: 
boosting SR performance metrics.

In recent years, CNN-based SR methods mainly improve performance by 
making networks deeper \cite{SR:VDSR, SR:RCAN}, 
refining inter-layer connections \cite{SR:RDN:Conf, SR:CMSC, SR:OISR},
or embedding enhanced self-attention \cite{SR:SAN, SR:CSNLN, SR:DRLN, SR:NLSN}.
These methods share a common feature:
they learn an implicit feature processing mode that 
uses the weighted sum of a feature map's own features 
to achieve reconstruction of high-frequency details.
For example, CNN conceals the high-frequency information 
of the training data within its convolution kernels, 
revealing the learned detailed features only after the 
convolution operation, which is
similar to the interpolation method like Bicubic.
Similarly, the non-local (self-attention) expands the 
local interpolation to the global and learns an adaptive 
weight to perform the weighted sum operation. 
Different from these modes that learn features implicitly, 
we observe that the early dictionary learning-based
methods \cite{SR:ANR, SR:Sparse_Coding:Conf, SR:Sparse_Coding:Journ} 
explicitly preserve features from training data,
abstracting the feature decompositions by optimizing a
LR-HR dictionary pair.
Based on the above analysis, we design to integrate the
dictionary learning into CNN to expand its 
representation mode, thus improving its
representation ability and SR performance.
We propose a Trainable Feature Matching (TFM) module.
Within TFM, trainable feature sets are included to explicitly learn
the features from training images through feature matching.
Differing from the dictionary learning-based methods that require 
complex sparse optimization, 
TFM evaluates similarities between input LR feature and the
preserved LR feature set as the weight to perform reconstruction.
The training parameters occupied by the feature sets are 
few, so TFM creates a condition for building a model with
a small number of parameters.
Experiments show that TFM not only enhances the SR performance, 
but also reduces the training parameters of the original CNN-based model.

To further enhance SR performance, we adopt non-local into our model.
However, its resource consumption is substantial, growing quadratically 
with the image size.
Some methods \cite{SR:NLRN, SR:SAN, SR:CSNLN} apply Region-Level 
Non-Local (RLNL) to reduce its resource usage.
RLNL recursively divides input into blocks smaller than a 
certain size and then performs non-local on each block.
However, due to the inconsistency of the divided sizes, RLNL 
can only perform non-local on each block serially, which is inefficient.
In addition, there are lightweight versions such as NLSA \cite{SR:NLSN}, 
which prunes attention blocks through Locality Sensitive Hashing (LSH).
Although NLSA works well, its implementation requires multiple 
LSH and tedious chunk size selection, which is complicated.
In this paper, we propose a simple lightweight version named
Same-size-divided Region-level Non-Local (SRNL).
SRNL simply divides the feature map into blocks with a uniform size,
so that all blocks can independently execute non-local operation in parallel. 
Experiments show that SRNL not only dramatically 
reduces the computation and memory consumption of the original non-local, 
but achieves even better SR performance.

In this paper, we propose Trainable Feature Matching Attention Network (TFMAN).
As shown in Figure \ref{fig_network}, recurrent convolutional network is 
adopted as the backbone. 
Three branches are included in the recurrent block.
We apply multiple projected fusion to fuse the three branches, 
which proved to be better than directly concatenating \cite{SR:CSNLN}.
The proposed TFM and SRNL are respectively located in two branches.
During implementation, through experiments, 
we found that TFM performs feature reconstruction independently 
on each channel better than simultaneously on all channels,
so we add a Channel Attention (CA) after the TFM 
to make model focus more on important channels.
The details will be explained in section \ref{TFM_Methodology},
and the experimental results will be shown in Table \ref{table_ablation_comparison}.

The contributions of this paper can be summarized as follows:

\begin{itemize}
\item
We propose a novel Trainable Feature Matching (TFM) module, which explicitly 
preserves training features to enhance the representation ability of CNNs for SR.
\item
We propose Same-size-divided Region-level Non-Local (SRNL) to reduce resource 
consumption and improve SR performance of the original non-local mechanism.
\item
Applying TFM, SRNL and Channel Attention (CA), and
taking recurrent convolutional network as the backbone,
we propose Trainable Feature Matching Attention Network (TFMAN).
Extensive experiments show that
TFMAN achieves better quantitative and qualitative results on most 
benchmark datasets with a smaller number of parameters
compared with other state-of-the-art methods.
  
\end{itemize}

\section{Related Works}
In this section, we first review the top performers and 
important modules used in CNN-based methods. 
Subsequently, concerning the newly proposed TFM and SRNL, methods that
explicitly preserve features and methods employing non-local are introduced. 
Finally, we retrospect methods adopting attention mechanism, as is 
used in our network.

\subsection{CNN-based SR Methods}
CNN were widely used in SR methods after \citeauthor{SR:SRCNN}
proposed SRCNN \cite{SR:SRCNN}. 
\citet{SR:VDSR} designed VDSR and demonstrated 
that the skip connection reduces the burden of the skipped module to 
pass the identity information.
By recursively reusing a same module multiple times, 
DRRN \cite{SR:DRRN} achieved outstanding performance with a small number
of training parameters. 
 \citet{SR:EDSR} showed that Batch Normalization (BN) in a residual block 
is harmful to SR, so they removed it and modified the residual 
connection to build EDSR.
Methods like \cite{SR:CARN, SR:RDN:Conf} % , J_PR_SR_HDRN
utilizes cascade connection or dense connection 
to achieve more efficient propagation of information between layers. 
To enhance the robustness of the model in dealing with various degradation,
SRMD \cite{SR:SRMD} accepts not only the LR image that should be restored, 
but also the blur kernel and noise level that degrade the original HR image. 
TDPN \cite{SR:TDPN} inputs the difference between an image and its 
blurred counterpart at training stage to enhance its texture restoring ability.
LapSR \cite{SR:LapSRN:Conf, SR:LapSRN:Journ} 
% and ProSR \cite{SR:ProSR} 
adopt the strategy of step-by-step training to design models that
progressively performs SR tasks with different scales from small to large, which 
alleviates the training difficulty of SR tasks with large scale factor like x8. 
% S$^2$TSR\cite{J_PR_SR_S2TSR} utilizes adversarial
% learning to transfer the knowledge from supervised learning to unsupervised one, 
% so as to learn more degradation modes from unlabeled images.
\citet{J_PR_SR_SISR_PF_OA} applies modules with 1D and 2D 
convolution kernels to extract orientation-aware features.
Additionally, there are some CNN-based methods that applies 
non-local and attention mechanism, which will be mentioned below.

\subsection{Methods Explicitly Preserving Features and Non-Local} 
Before the popularity of CNN, dictionary learning was the mainstream 
to implement image SR, such as 
 \cite{SR:ANR, SR:Sparse_Coding:Conf, SR:Sparse_Coding:Journ}, 
which optimize an LR-HR dictionary pair to preserve the feature 
decompositions of training image patches.
They are typical methods that explicitly preserve training
features. However, limited by the size of the dictionary pair, their 
SR performance are hard to catch up with CNN-based methods, which 
possess much more training parameters. 
Therefore, we propose Trainable Feature Matching (TFM) module and
embed it into CNN to combine the advantages of both.

Recently, \citet{Non_local} proposed non-local neural networks, which 
generalized self-attention as non-local operation to 
build long-range feature correlations in video sequence. 
Combined with non-local, RNAN \cite{SR:RNAN} and NLRN \cite{SR:NLRN} 
respectively implement SR on a residual network and a recurrent network.
\citet{SR:CSNLN} expand non-local to Cross-Scale Non-Local (CS-NL)
to exploit cross-scale self-similarity existing in images.
However, non-local is computationally complex and memory consuming, 
which is quadratic to image size.
Therefore, its lightweight versions have been proposed successively.
SAN \cite{SR:SAN} executes non-local on blocks that is recursively divided 
smaller than a certain size.
However, its efficiency is not that high because of serial execution of 
non-local on each block with inconsistent size.
\citet{Restormer} implement attention between channel dimension
rather than original spatial dimension to reduce the computation of 
non-local.
However, this changes the nature of non-local, 
which can not make the model pay attention to the relationships 
between spatial pixels.
NLSA \cite{SR:NLSN} 
divide similar positions into the same block that require attention
through spherical Locality Sensitive Hashing (LSH),
so as to avoid matching noise in non-local.
However, the number of pixels in each bucket is not necessarily same after LSH.
To solve this problem, the authors evenly divide the pixels sorted
by the bucket number into chunks and do attention on these chunks.
Then, these operation will be executed multiple times 
and their average results will be taken to mitigate the
disadvantages coming from cross bucket attention.
Different from the above, 
we propose a simple and effective version, 
Same-size-divided Region-level Non-Local (SRNL). 
SRNL not only retains the parallelism and the
original spatial characteristic, 
but has no complicated multiple hashes 
and tedious chunk size selection like NLSA \cite{SR:NLSN}.

\subsection{Attention Mechanism}
Attention mechanism \cite{Conv_attention} is used to enhance or 
weaken features along a certain dimension to make the model focus 
more on important features, such as Channel Attention (CA) 
and Spatial Attention (SA).
They are widely used in SR methods 
\cite{SR:RCAN, SR:DRLN, SR:CSFM, J_PR_SR_CASGCN}.
Furthermore, there are several improved attention mechanisms. 
For instance, 
\citet{SR:SAN} replaced the prior of the original CA with the 
covariance matrix between channels of the input feature map
to better carry the relationships between channels.
Similarly, \citet{J_PR_SR_DiVANet} extended the second-order attention 
along the horizontal and vertical dimensions.
\citet{SR:RFANet} designed a lightweight SA as a foundational element
for extremely deep networks.

\section{Proposed method}
This section includes five subsections. The first two subsections
introduce the overall structure of TFMAN.
In the remaining three subsections, each module adopted in our model will be 
introduced.

\begin{figure*}[!t]
    \centering
    \includegraphics[width=\linewidth]{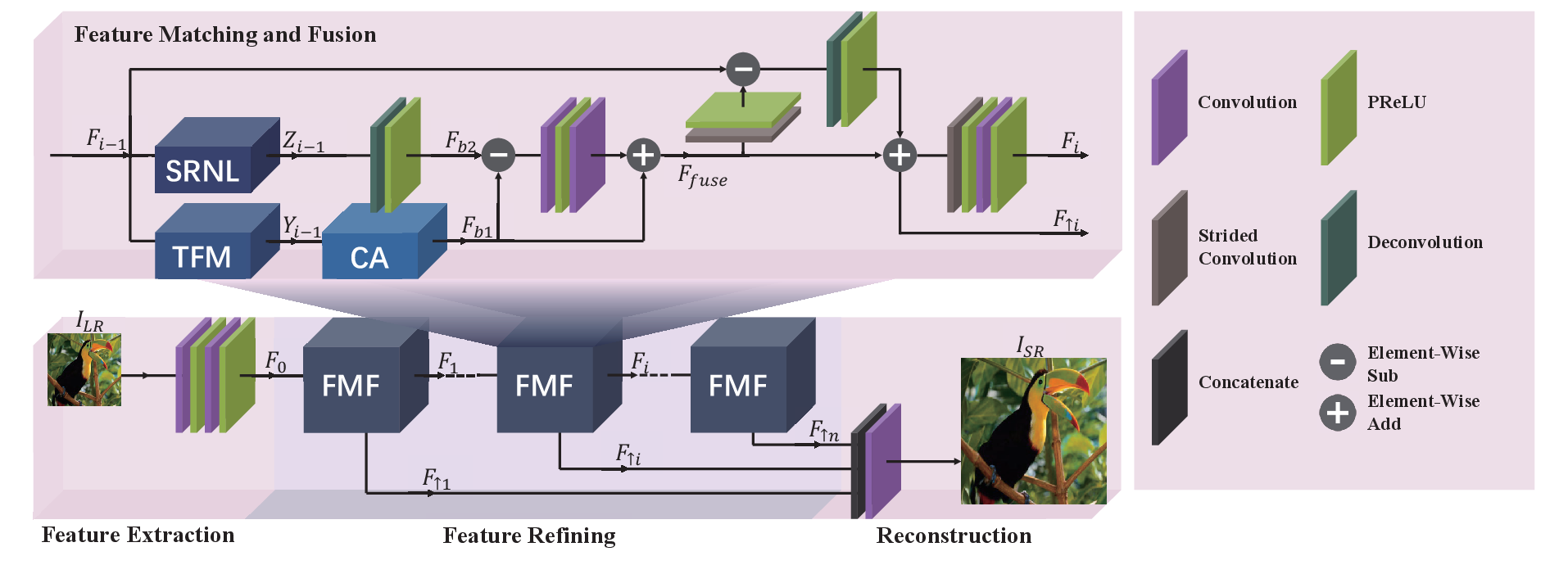} 
    \caption{
      Architecture of proposed TFMAN. 
                              }
    \label{fig_network}
    \end{figure*}
\subsection{Network Framework}
As displayed in the lower part of Figure \ref{fig_network}, the proposed 
Trainable Feature Matching Attention Network (TFMAN) consists of three parts: 
feature extraction part, feature refining part, and reconstruction part. 
Following \cite{SR:NLRN, SR:CSNLN}, 
we apply the recurrent network that fuses all recurrent outputs at the tail as our model's backbone.
Assuming that $I_{LR}\in R^{3\times H\times W}$ 
and $I_{SR}\in R^{3\times sH\times s W}$ are the input and output of our model,
the model first extracts feature $ F_0\in R^{C\times H\times W}$ through
\begin{equation}
  F_0=H_{FE}\left(I_{LR}\right),
\end{equation}
where $H_{FE}$ contains two convolutions that map shallow feature to middle 
feature. For simplicity, the activation function is omitted. Unless specified, we 
apply Parametric Rectified Linear Unit (PReLU) \cite{PReLU} throughout the network. 
After that, the extracted feature $F_0$ is fed into feature refining part and 
carried out the following recurrent operations. 
Every recurrence, the Feature Matching and Fusion (FMF) module
outputs an up-sampled feature map
$F_{\uparrow i}\in R^{C\times sH\times s W}$ 
for the final global feature fusion:
\begin{equation}
\begin{aligned}
  F_{\uparrow i}=FMF_i^{\prime\prime}\left(FMF_{i-1}^\prime\left(\ldots F M F_1^\prime
  \left(F_0\right)\right)\right),&\\
  i=1,2,\ldots,n,&
\end{aligned}
\end{equation}
where $FMF_i^\prime, FMF_i^{\prime\prime}$ denote the branch whose output should 
be fed into the next recurrence and the branch whose output is used for the final 
feature fusion, at $i$-th recurrence. 

Finally, in reconstruction part, all the features up-sampled from $n$ recurrences 
are channel-wisely concatenated and then mapped to $3$ channels through a 
convolution to obtain the super-resolution result, i.e., 
\begin{equation}
  I_{SR}=H_{Rec}\left(F_{\uparrow1},F_{\uparrow2},\ldots,F_{\uparrow n}\right).
\end{equation}

For training, TFMAN is optimized with $L_1$ loss. Given a training set 
$\{I_{LR}^i,I_{HR}^i\}_{i=1}^M$ that contain $M$ LR-HR image pairs. 
The optimization objective is defined as (TFMAN is abbreviated as $TN$)
\begin{equation}
  TN^\ast = \arg\min_{TN}\frac{1}{M}\sum_{i=1}^M\left\|TN\left(I^i_{LR}\right)-I^i_{HR}\right\|_1.
\end{equation}

\subsection{Feature Matching and Fusion (FMF)}
As shown in the upper part of Figure \ref{fig_network}, FMF 
gathers three branches: identity mapping branch, SRNL branch and TFM branch.
Among them, the TFM, SRNL and CA module will be discussed in later subsections.  
The common practice to connect branches is channel-wisely concatenating 
feature maps at once, or simply adding them together,
e.g. \cite{SR:EDSR, SR:RNAN, SR:NLRN}. 
However, through comparative experiments, \citet{SR:CSNLN} discovered that 
gradually fusing branches by handling difference between branch outputs  
is more beneficial, so it is adopted in FMF, which can be simply defined as
\begin{equation}
  f=OPs\left(b_1-b_2\right)+b_2,
\end{equation}
where $b_1,b_2,f, OPs$ sequentially indicate the outputs of two branches, the 
fusion result and some operations to handle the difference. 
Given $F_{i-1}\in R^{C\times H\times W}$ as the input of FMF at $i$-th 
recurrence. Before the first fusion, we obtain
\begin{align}
  &F_{b1}=C\!A\left(T\!F\!M\left(F_{i-1}\right)\right),\\
  &F_{b2}=C_{\uparrow1}\left(S\!R\!N\!L\left(F_{i-1}\right)\right),
\end{align}
where $C_{\uparrow1}, F_{b1}\in R^{C\times sH\times sW}, F_{b2}\in R^{C\times sH\times sW}$
respectively denote a deconvolution, the output of TFM branch and the output of 
SRNL branch. Then we obtain the first fusion result
\begin{equation}
  F_{fuse}=OPs_1\left(F_{b2}-F_{b1}\right)+F_{b1},
\end{equation}
where $OPs_1$ contains two convolutions. After that, the identity 
mapping branch is also fused in through
\begin{equation}
  F_{\uparrow i}=C_{\uparrow2}
  \left(F_{i-1}-C_{\downarrow1}\left(F_{fuse}\right)\right)+F_{fuse},
\end{equation}
where $C_{\downarrow1}$ and $C_{\uparrow2}$ denote a strided convolution and a 
deconvolution. Here, $F_{\uparrow i}\in R^{C\times sH\times s W}$ is the up-sampled 
feature at $i$-th recurrence for the final global feature fusion. 
Afterwards, it will be down-sampled back to the space of 
middle feature map to get the input of next recurrence:
\begin{equation}
  F_i=C_{\downarrow2}\left(F_{\uparrow i}\right),
\end{equation}
where $C_{\downarrow2}$ contains a scale strided convolution and a
normal convolution.

\begin{figure*}
    \centering
    \begin{minipage}[b]{0.36\linewidth}
        \centering
        \subfloat[TFM]{
          \includegraphics[width=0.95\linewidth]{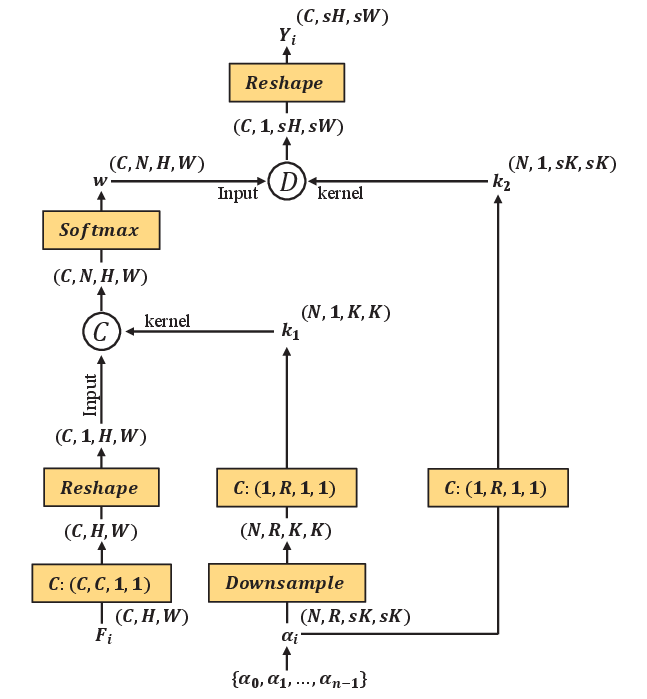}
          \label{fig_TFM}
        }
    \end{minipage}  
    \begin{minipage}[b]{0.63\linewidth}
        \centering
        \subfloat[SRNL]{
          \includegraphics[width=0.95\linewidth]{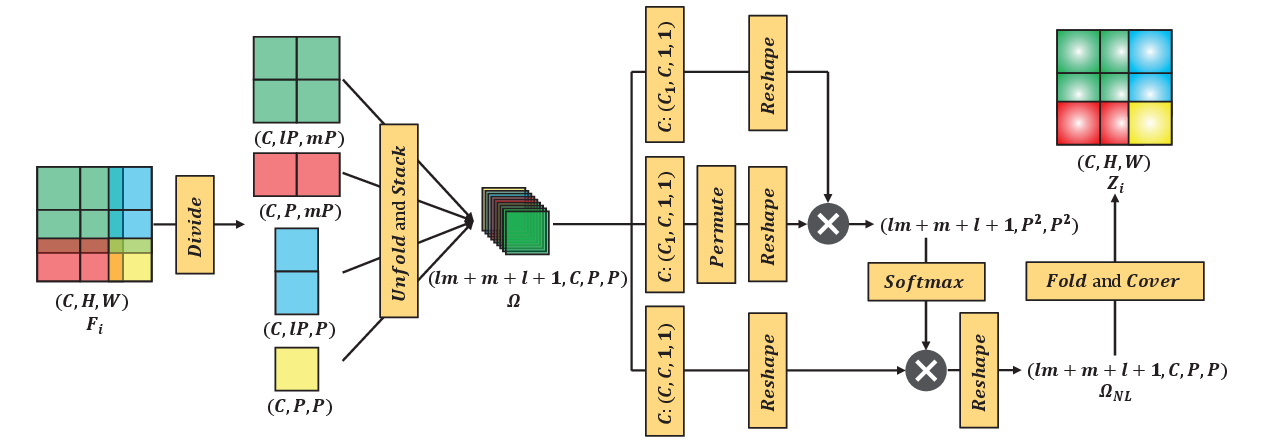}
          \label{fig_SRNL}
        }\\
        \subfloat[CA]{
          \includegraphics[width=0.95\linewidth]{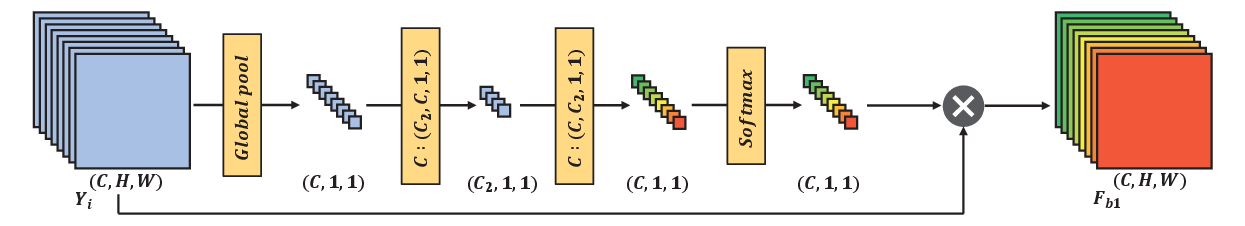}
          \label{fig_CA}
        }
    \end{minipage}
    \caption{
      Illustrations of TFM, SRNL and CA modules. 
      In sub-figure (a), the $C$ and $D$ around circle
      respectively represent convolution and deconvolution. 
      In sub-figure (b), the cross around circle represents matrix 
      multiplication, but in the sub-figure (c), it represents element-wise
      multiplication. Additionally, among the operations enclosed by yellow 
      box, $C\!\!:\!(C_1,C_2,K,K)$ represents the convolution mapping 
      features from $C_2$ channels to $C_1$ channels 
      with $K\!\times\!K$ sized convolution kernel.
    } 
    \label{fig_modules}
  \end{figure*} 
\subsection{Trainable Feature Matching (TFM) Module}
\label{TFM_Methodology}
For CNN, the convolution once performs a weighted sum on a neighborhood,
which is similar to interpolation methods like Bicubic.  
Therefore, what the convolution kernel learned can be regarded as a specific 
interpolation mode, which implicitly preserves the features of training data.
In order to enhance the information retention capacity of the model to the 
training data, designing new information retention modes is a favorable idea.
Here, we propose Trainable Feature Matching (TFM) module, 
which explicitly learns the features from the training data, 
to add a new form of representation learning to the CNN-based model.
The early methods based on sparse coding
\cite{SR:ANR, SR:Sparse_Coding:Conf, SR:Sparse_Coding:Journ}
are typical methods that explicitly preserve the training features. 
Through sparse optimization, they extract feature decompositions from
training image patches.
However, in order to constitute a module utilized in CNN-based model, 
it is necessary to avoid this complex optimization.
The method based on sparse coding sparsely optimizes the similarity between 
the linear combination of the feature decompositions in LR dictionary and 
the patch to be reconstructed.
In contrast, TFM directly matches the similarity between the down-sampled 
preserved features and the patch to be reconstructed. The more similar 
the features to the patch, the higher the weight of the features will be gotten.
Finally, these weights will be used for the weighted sum of the features to 
complete the reconstruction of the patch.
Below we first introduce the basic idea of TFM module, and then detail its 
implementation in proposed TFMAN.

As shown in Figure \ref{fig_TFM}, $n$ feature sets 
$\{\alpha_1,\alpha_2,\ldots,\alpha_n\}$ with shape of $(N,R,sK,sK)$ 
are defined for feature matching of $n$ recurrences,  
where $N,R,s$ and $K$ are the feature number, feature channel number, 
scale factor and feature size. 
These $nNRs^2K^2$ parameters of the feature sets are trainable and will be 
updated with gradient descent algorithm during training.
Given a patch $f$ with shape of $(R^\prime,K,K)$ that needs to be reconstructed,
and the feature set $\alpha$ used in this reconstruction,
the basic idea of TFM is defined as follows:
\begin{equation}\label{TFM_define}
  f_\uparrow=\sum_{i}^{N}{\frac{\exp{\left(\Phi\left(f,\Lambda(\alpha^i)\right)\right)}}
  {\sum_{j}^{N}\exp{\left(\Phi\left(f,\Lambda(\alpha^j)\right)\right)}}\Psi\left(\alpha^i\right)},
\end{equation}
where $\Phi,\Psi,\Lambda$ and $f_\uparrow$ are respectively a function to measure 
similarity, a transformation function, the operation to obtain down-sampled version 
of $\alpha$ with $R^\prime$ channels to match $f$, 
and the patch $f$ after reconstruction. 
As shown in equation \ref{TFM_define}, first,
features from $\alpha$ are down-sampled and mapped to the space of $f$
through a certain interpolation method.
Then, these down-sampled features are matched with $f$ and measure 
similarity scores between them.
Finally, the patch $f$ is reconstructed through a 
weighted summation of features derived from $\alpha$, 
wherein normalized similarity scores are employed 
as the corresponding weights.

Below we elaborate the implementation of TFM in TFMAN, which is clearly
shown in \ref{fig_TFM}. 
Given an input $F_i\in R^{C\times H\times W}$ at $i$-th recurrence. 
It is first transformed in preparation for feature matching:
\begin{equation}
  F_{match}=Reshape\left(H_{FT}\left(F_i\right)\right),
\end{equation}
where $H_{FT}$ is a $1\times1$ convolution with identical channel mapping 
and $F_{match}$ is shaped as $(C, 1, H, W)$. 
The reason for using reshape operation is that the feature matching and 
reconstruction are implemented on each independent channel, i.e. $R^\prime=1$. 
Therefore, the $f$ in equation \ref{TFM_define} can be represented as a patch
with shape of $(1,K,K)$ extracted from $F_{match}$ of certain channel.
Meanwhile, the feature set 
$\alpha_i$
is also transformed to get matching samples:
\begin{equation}
  k_1=H_{\alpha T1}\left(\alpha_i\right),
\end{equation}
where $H_{\alpha T1}$ is corresponding to $\Lambda$ in equation \ref{TFM_define},
which contains an interpolation down-sampling and a $1\times1$ 
convolution mapping $R$ channels to $1$ channel, 
so that the matching samples $k_1$ is with shape of $(N,1,K,K)$. 
After that, the feature matching $\Phi$ is implemented
as convolution whose kernel is $k_1$ and input is $F_{match}$. Then, the output 
will go through a softmax function along the second order to measure how similar 
are the patches of each channel to the features. 
These can be formulated as
\begin{equation}
  w=Softmax\left(Conv\left(k_1,F_{match}\right)\right).
\end{equation} 
As such, $w\in R^{C\times N\times H\times W}$ is the summation weights 
of the reconstruction.
Before reconstruction, the reconstruction samples are obtained from
\begin{equation}
  k_2=H_{\alpha T2}\left(\alpha_i\right),
\end{equation}
where $H_{\alpha T2}$ is corresponding to $\Psi$ in equation \ref{TFM_define}, 
which contains a $1\times1$ convolution that maps the channel number 
from $R$ to $1$, so that the reconstruction 
samples $k_2$ is with shape of $(N, 1, sK, sK)$.
Finally, the reconstruction process is implemented as a deconvolution 
whose kernel is $k_2$ and input is $w$, with stride of $s$:
\begin{equation}
  Y_i=Reshape\left(Deconv\left(k_2,w\right)\right),
\end{equation}
where the reshape operation shapes the output from 
$(C,1,sH,sW)$ to $(C,sH,sW)$.

Now we give explanations on two details:

1) In the implementation of the TFM, due to the overlapping of matched patches, 
a patch of the result $Y_i$ represents the average of the reconstructed patches 
that are adjacent in a $K\times K$ sized block.

2) As for feature matching implemented on each independent channel, 
we have also attempted to match all channels simultaneously. 
Experiments (see Table \ref{table_ablation_comparison}) 
show that its performance is not better than the former, 
and even takes up more parameters.

\subsection{Same-size-divided Region-level Non-Local (SRNL)}
Non-local attention has recently played a significant role in computer vision. 
However, due to the necessity of computing a relationship matrix among all 
pixels, non-local requires substantial computational and memory resources.
To address this, we propose Same-size-divided Region-level Non-Local (SRNL), 
in which non-local attention is independently applied to each block 
with a fixed size. SRNL not only significantly reduces the 
computational and memory requirements compared to the original non-local,
but it also slightly improves the performance of SR task,
as verified in Figure \ref{fig_SRNL_patch_size_comp}.

As shown in Figure \ref{fig_SRNL}, given input feature map $F_i$ with shape of
$(C,H,W)$, SRNL first divide $F_i$ into four large overlapping 
blocks respectively with size of 
$\left(lP, mP\right)$, $\left(P, mP\right)$, 
$\left(lP, P\right)$ and $\left(P, P\right)$, 
so that each large blocks can be completely divided into several 
non-overlapping small blocks with size of $\left(P,P\right)$. 
Then, all these small blocks are stacked as $\Omega$ 
with shape of $\left(lm\!+\!m\!+\!l\!+\!1, C, P, P\right)$ for performing 
independent non-local on each small blocks. 
Above can be formulated as
\begin{equation}
  \Omega=Stack\left(Unfold\left(Divide\left(F_i\right)\right)\right),
\end{equation}
where the second division can be realized by $Unfold$ function in Pytorch.
Afterwards, non-local is independently performed on these $lm\!+\!m\!+\!l\!+\!1$ 
small blocks, which can be formulated as 
\begin{equation}
\begin{aligned}
  \Omega_{NL}^{t,:,a,b}=\sum_{c,d}\frac
  {\exp{\left(\Phi\left(\Omega^{t,:,a,b},\Omega^{t,:,c,d}\right)\right)}}
  {\sum_{e,f} \exp{\left(\Phi\left(\Omega^{t,:,a,b},\Omega^{t,:,e,f}\right)\right)}}
  \Psi\left(\Omega^{t,:,c,d}\right),&\\
  t = 1,2,...,lm\!+\!m\!+\!l\!+\!1,&\\
  a,b,c,d,e,f = 1,2,...,P,&
\end{aligned}
\end{equation}
where $\Omega^{t,:,i,j},\Phi,\Psi$ respectively represent a feature in \mbox{$t$-th}
block at position of $\left(i,j\right)$, a similarity function, 
and a transformation function. 
As shown in Figure \ref{fig_SRNL}, matrix multiplication and 
$1\times1$ convolutions are adopted respectively to measure 
similarity and transform features,
which is conventional practice of non-local,
so we will not elaborate it again. 
After non-local, stacked small blocks in 
$\Omega_{NL}$ are
reversely merged to form large blocks with aforementioned four sizes, i.e. 
$\left(lP,mP\right)$, $\left(P,mP\right)$,
$\left(lP,P\right)$ and $\left(P,P\right) $, 
and then these large blocks are covered to the original positions of 
the feature map to get the output of SRNL:
\begin{equation}
  Z_i=Cover\left(Fold\left(\Omega_{NL}\right)\right),
\end{equation}
where the merging operation can be realized by $Fold$ function in Pytorch. Note 
that, if the height or width of the input $F_i$ is not divisible by the width of 
the small block, i.e. $\left(H,W\right)\mod{P}\neq\left(0,0\right)$, some 
features will be covered by others, as illustrated in Figure \ref{fig_SRNL}.

Below, we compare the computation between SRNL and the original non-local.
For the convenience of calculation, 
we set the number of converted channels $C_1=C$.
The amount of Multiply ACcumulate (MAC) operations of 
the two can be respectively expressed as follows:
\begin{equation}
\begin{aligned}
  &\begin{aligned}
    MAC_{NL} &= 3MAC_{Trans}+MAC_{Mactch}+MAC_{Rec}\\
    &= 3WHC^2 + 2H^2W^2C,\\ 
  \end{aligned}\\
  &\begin{aligned}
    MAC_{SRNL} 
        &= 3\left\lceil\frac{W}{P}\right\rceil \left\lceil\frac{H}{P}\right\rceil P^2C^2 + 
       2\left\lceil\frac{W}{P}\right\rceil \left\lceil\frac{H}{P}\right\rceil P^4C.
      \end{aligned}
\end{aligned}
\end{equation}
From the formula, we can conclude that the larger the size of the input feature 
map, the larger the acceleration ratio of SRNL relative to non-local.
Supposing that $C = 128, P = 48$ and the input size is $640 \times 360$, 
it can be calculated that the computation of non-local operation is 83 
times that of SRNL.
Furthermore, we can also calculate that the memory peak of the non-local is 
$640^2*360^2*4B=197.7GB$, but it is only $14*8*48^4*4B = 2.2GB$ for SRNL.

\subsection{Channel Attention (CA)}
\label{CA_Methodology}
Attention mechanisms are widely used in SR models 
\cite{SR:RCAN,SR:CSFM,SR:SAN,SR:RFANet}. 
Since TFM performs feature matching and reconstruction on each channel 
independently, we connect a Channel Attention (CA)
after TFM to enhance correlation between channels.
The effectiveness of this strategy is proved by experiments 
(see Table \ref{table_ablation_comparison}).
As shown in Figure \ref{fig_CA}, 
given input $Y_i$, CA can be shortly defined as
\begin{equation}
  F_{b1}=Softmax\left(Conv\left(Conv\left(H_{GP}\left(Y_i\right)\right)\right)\right)
  \times Y_i,
\end{equation}
where $H_{GP}$ is spatial global average pooling. The two convolutions are with 
$1\times1$ sized kernels, which first maps the feature 
(with size of $(1,1)$) to a smaller space and then maps it back. 
For the final multiplication operation, the attention value is first 
broadcasted along the spatial dimension to the same size as the input $Y_i$, 
and then element-wisely multiply with $Y_i$.

\section{Experiments}

In this section, we begin by introducing the settings used during the training 
and testing of our model. Afterwards, quantitative and qualitative comparisons 
between TFMAN and state-of-the-art methods are conducted with 
Bicubic (BI), Blur-Downscale (BD), and Downscale-Noise (DN) degradation models.
Following this, we delve into the overall architecture of our model. 
Finally, we present our investigations into 
the newly proposed TFM and SRNL.

\subsection{Settings}
\subsubsection{Datasets and Evaluation Metrics}
Following \cite{SR:EDSR, SR:RCAN, SR:RNAN, SR:SAN}, 
our model is trained on the first 800 images of DIV2K training dataset. 
For testing, benchmark datasets Set5, Set14, BSD100, Urban100 and Manga109, 
are adopted. For evaluation, Peak Signal-to-Noise Ratio (PSNR) and 
Structural SIMilarity (SSIM) metrics are calculated on Y channel 
of transformed YCrCb space for all SR results. 

\subsubsection{Degradation Models}
To comprehensively showcase the superiority of proposed TFMAN, we employ 
three degradation models to generate LR counterparts from HR images. 
These models encompass Bicubic (BI), 
Blur-Downscale (BD), and Downscale-Noise (DN).
BI is the basic degradation model that implements Bicubic down-sampling by 
calling $resize$ function in Matlab with the Bicubic option. 
BD first blurs the HR image by $7\times 7$ sized Gaussian kernel with
standard deviation 1.6, and then performs BI degradation. 
DN first performs BI degradation and then adds Gaussian noise with noise 
level 30 into the image. Since DN imports a lot of irrelevant information 
into LR images, it is the most difficult to recover among the three 
degradation models. For the scale factors of experiments, with BI degradation, 
LR images are generated with scale factors of 2, 3, 4 and 8. 
For BD and DN degradation models, following \cite{SR:SRFBN},
LR images with scale factor of 3 are generated.

\subsubsection{Model Setting} 
Overall, we set the recurrence number of FMF as $12$ and the
channel number of middle features as 128, i.e. $n=12,C=128$. 
Except for those that have been specified as $1\times1$ in modules, 
all other convolution kernels' sizes are set as $3\times3$. 
In FMF, for up-sampling, kernel size of deconvolutions are 
set as $3s\times 3s$, where $s$ represents scale factor. 
For down-sampling, when scale factor is powers of $2$, we divide it into $t$ 
steps. Each step contains one strided convolution to perform a double 
down-sampling. Therefore, we set the kernel size of strided convolutions as $6\times6$
for scale factor of $2^t$, and $9\times9$ for scale factor of $3$.
In TFM, the size of trainable features are set as $3s\times3s$, i.e. $K=3$. 
In order to balance performance and computation, we set the number of feature
$N=32$ and the number of feature channel $R=4$.
As for the down-sampling of the feature sets, bilinear is adopted. 
In SRNL, we set the small block’s size $P=48$ and the transformed 
channel number $C_1=64$. 
In CA, we set reduced channel number $C_2=8$.

\subsubsection{Training Details} 
For each iteration, $16$ LR-HR image pairs are randomly sampled from training 
set and randomly cropped into patches with size of $48\times48$. 
Before feeding into the model, the patches are augmented by randomly rotating 
$0^\circ, {90}^\circ, {180}^\circ, {270}^\circ$ and mirror flipping.
The model is experimented on RGB color space.
For optimization, we apply ADAM optimizer with 
$\beta_1=0.9, \beta_2=0.999$, and $\epsilon=5\text{e}-7$. 
We train 15000 epochs from scratch for BI degradation model,
where one epoch contains $50$ iterations.
For BD and DN degradation models, we train 8500 epochs from the
pretrained BI model.
The learning rate is initialized as ${10}^{-4}$, and halved at 5000th, 8500th, 
10500th, 11500th, 12500th, 13500th epoch for BI, and at 2000th, 3500th, 4500th, 
5500th, 6500th, 7500th epoch for BD and DN.
We apply `O2' mode of Apex to accelerate the training, 
which is a mixed precision training module.
The proposed model is implemented on Pytorch and trained on an Nvidia RTX 3090 GPU.

\subsection{Comparisons with State-of-the-Art Methods}

% Review 3
\begin{table*}[!t]
    \renewcommand{\arraystretch}{0.6} % 表示几倍的行高
    \caption{Quantitative comparison with BI degradation model for 
    x2, x3, x4 and x8. 
    The red and the blue index respectively denote the best and 
    the second best results.} 
    \label{table_compare_BI}
    \centering
    \tiny
    
    \begin{tabular}{|*{15}{@{}c@{}|}} %表格格式在此定义
      \hline
      \multirow{2}{*}{Scale}&\multirow{2}{*}{Method}&
      \multicolumn{2}{c|}{Set5}&\multicolumn{2}{c|}{Set14}&\multicolumn{2}{c|}{BSD100}&
      \multicolumn{2}{c|}{Urban100}&\multicolumn{2}{c|}{Manga109}&\multicolumn{2}{c|}{Average}\\
      \cline{3-14}
      &&PSNR&SSIM&PSNR&SSIM&PSNR&SSIM&PSNR&SSIM&PSNR&SSIM&PSNR&SSIM\\
      \hline
      \multirow{14}{*}{x2} 
      & RCAN \cite{SR:RCAN} & 38.27  & 0.9614  & 34.11  & 0.9216  & 32.41  & 0.9026  & 33.34  & 0.9384  & 39.43  & 0.9786  & 35.51  & 0.9405  \\
      & OISR-RK3 \cite{SR:OISR} & 38.21  & 0.9612  & 33.94  & 0.9206  & 32.36  & 0.9019  & 33.03  & 0.9365  & - & - & 34.39  & 0.9301  \\
      & CSFM \cite{SR:CSFM} & 38.26  & 0.9615  & 34.07  & 0.9213  & 32.37  & 0.9021  & 33.12  & 0.9366  & 39.40  & 0.9785  & 35.44  & 0.9400  \\
      & RFANet \cite{SR:RFANet} & 38.26  & 0.9615  & 34.16  & 0.9220  & 32.41  & 0.9026  & 33.33  & 0.9389  & 39.44  & 0.9783  & 35.52  & 0.9407  \\
      & SAN \cite{SR:SAN} & 38.31  & 0.9620  & 34.07  & 0.9213  & 32.42  & 0.9028  & 33.10  & 0.9370  & 39.32  & 0.9792  & 35.44  & 0.9405  \\
      & CASGCN \cite{J_PR_SR_CASGCN} & 38.26  & 0.9615  & 34.02  & 0.9213  & 32.36  & 0.9020  & 33.17  & 0.9377  & 39.41  & 0.9785  & 35.44  & 0.9402  \\
      & CSNLN \cite{SR:CSNLN} & 38.28  & 0.9616  & 34.12  & 0.9223  & 32.40  & 0.9024  & 33.25  & 0.9386  & 39.37  & 0.9785  & 35.48  & 0.9407  \\
      & SwinIR \cite{SwinIR} &\textcolor{blue}{38.35}  & 0.9620  & 34.14  & 0.9227  & 32.44  & 0.9030  & 33.40  & 0.9393  &\textcolor{blue}{39.60}  & 0.9792  & 35.59  & 0.9412  \\
      & NLSN \cite{SR:NLSN} & 38.34  & 0.9618  & 34.08  & 0.9231  & 32.43  & 0.9027  & 33.42  & 0.9394  & 39.59  & 0.9789  & 35.57  & 0.9412  \\
      & DRLN \cite{SR:DRLN} & 38.27  & 0.9616  &\textcolor{red}{34.28}  & \textcolor{blue}{0.9231}  & 32.44  & 0.9028  & 33.37  & 0.9390  & 39.58  & 0.9786  & 35.59  & 0.9410  \\
      & Restormer \cite{Restormer} & 38.15  & 0.9609  & 33.36  & 0.9140  & 32.27  & 0.9007  & 31.33  & 0.9171  & 33.46  & 0.9552  & 33.72  & 0.9296  \\
      & CRAN \cite{SR:CRAN} & 38.31  & 0.9617  & 34.22  &\textcolor{red}{0.9232}  & 32.44  & 0.9029  &\textcolor{blue}{33.43}  &\textcolor{blue}{0.9394}  &\textcolor{red}{39.75}  &\textcolor{blue}{0.9793}  &\textcolor{blue}{35.63}  & 0.9413  \\
      & TDPN \cite{SR:TDPN} & 38.31  &\textcolor{blue}{0.9621}  & 34.16  & 0.9225  &\textcolor{red}{32.52}  &\textcolor{red}{0.9045}  & 33.36  & 0.9386  & 39.57  &\textcolor{red}{0.9795}  & 35.58  &\textcolor{blue}{0.9414}  \\
      & TFMAN(ours) &\textcolor{red}{38.36}  &\textcolor{red}{0.9621}  &\textcolor{blue}{34.27}  &0.9228  &\textcolor{blue}{32.46}  &\textcolor{blue}{0.9031}  &\textcolor{red}{33.63}  &\textcolor{red}{0.9410}  & 39.58  & 0.9789  &\textcolor{red}{35.66}  &\textcolor{red}{0.9416}  \\

      \hline
      \multirow{14}{*}{x3}   
      & RCAN \cite{SR:RCAN} & 34.74  & 0.9299  & 30.64  & 0.8481  & 29.32  & 0.8111  & 29.08  & 0.8702  & 34.43  & 0.9498  & 31.64  & 0.8818  \\
      & OISR-RK3 \cite{SR:OISR} & 34.72  & 0.9297  & 30.57  & 0.8470  & 29.29  & 0.8103  & 28.95  & 0.8680  & - & - & 30.88  & 0.8638  \\
      & CSFM \cite{SR:CSFM} & 34.76  & 0.9301  & 30.63  & 0.8477  & 29.30  & 0.8105  & 28.98  & 0.8681  & 34.52  & 0.9502  & 31.64  & 0.8813  \\
      & RFANet \cite{SR:RFANet} & 34.79  & 0.9300  & 30.67  & 0.8487  & 29.34  & 0.8115  & 29.15  & 0.8720  & 34.59  & 0.9506  & 31.71  & 0.8826  \\
      & SAN \cite{SR:SAN} & 34.75  & 0.9300  & 30.59  & 0.8476  & 29.33  & 0.8112  & 28.93  & 0.8671  & 34.30  & 0.9494  & 31.58  & 0.8811  \\
      & CASGCN \cite{J_PR_SR_CASGCN} & 34.75  & 0.9300  & 30.59  & 0.8476  & 29.33  & 0.8114  & 28.93  & 0.8671  & 34.36  & 0.9494  & 31.59  & 0.8811  \\
      & CSNLN \cite{SR:CSNLN} & 34.74  & 0.9300  & 30.66  & 0.8482  & 29.33  & 0.8105  & 29.13  & 0.8712  & 34.45  & 0.9502  & 31.66  & 0.8820  \\
      & SwinIR \cite{SwinIR} &\textcolor{red}{34.89}  &\textcolor{red}{0.9312}  & 30.77  &\textcolor{blue}{0.8503}  & 29.37  & 0.8124  & 29.29  & 0.8744  & 34.74  &\textcolor{red}{0.9518}  & 31.81  &\textcolor{blue}{0.8840}  \\
      & NLSN\cite{SR:NLSN} & 34.85  & 0.9306  & 30.70  & 0.8485  & 29.34  & 0.8117  & 29.25  & 0.8726  & 34.57  & 0.9508  & 31.74  & 0.8828  \\
      & DRLN \cite{SR:DRLN} & 34.78  & 0.9303  & 30.73  & 0.8488  & 29.36  & 0.8117  & 29.21  & 0.8722  & 34.71  & 0.9509  & 31.76  & 0.8828  \\
      & Restormer \cite{Restormer} & 34.63  & 0.9289  & 30.53  & 0.8454  & 29.24  & 0.8087  & 28.67  & 0.8608  & 34.16  & 0.9480  & 31.45  & 0.8784  \\
      & CRAN \cite{SR:CRAN} & 34.80  & 0.9304  & 30.73  & 0.8498  & 29.38  & 0.8124  &\textcolor{blue}{29.33}  &\textcolor{blue}{0.8745}  &\textcolor{red}{34.84}  & 0.9515  &\textcolor{blue}{31.82}  & 0.8837  \\
      & TDPN \cite{SR:TDPN} &\textcolor{blue}{34.86}  &\textcolor{blue}{0.9312}  &\textcolor{blue}{30.79}  & 0.8501  &\textcolor{red}{29.45}  &\textcolor{blue}{0.8126}  & 29.26  & 0.8724  & 34.48  & 0.9508  & 31.77  & 0.8834  \\
      & TFMAN(ours) & 34.84  & 0.9308  &\textcolor{red}{30.80}  &\textcolor{red}{0.8503}  &\textcolor{blue}{29.39}  &\textcolor{red}{0.8130}  &\textcolor{red}{29.41}  &\textcolor{red}{0.8764}  &\textcolor{blue}{34.77}  &\textcolor{blue}{0.9516}  &\textcolor{red}{31.84}  &\textcolor{red}{0.8844}  \\

      \hline
      \multirow{14}{*}{x4} 
      & RCAN \cite{SR:RCAN} & 32.62  & 0.9001  & 28.86  & 0.7888  & 27.76  & 0.7435  & 26.82  & 0.8087  & 31.21  & 0.9172  & 29.45  & 0.8317  \\
      & OISR-RK3 \cite{SR:OISR} & 32.53  & 0.8992  & 28.86  & 0.7878  & 27.75  & 0.7428  & 26.79  & 0.8068  & - & - & 28.98  & 0.8092  \\
      & CSFM \cite{SR:CSFM} & 32.61  & 0.9000  & 28.87  & 0.7886  & 27.76  & 0.7432  & 26.78  & 0.8065  & 31.32  & 0.9183  & 29.47  & 0.8313  \\
      & RFANet \cite{SR:RFANet} & 32.66  & 0.9004  & 28.88  & 0.7894  & 27.79  & 0.7442  & 26.92  & 0.8112  & 31.41  & 0.9187  & 29.53  & 0.8328  \\
      & SAN \cite{SR:SAN} & 32.64  & 0.9003  & 28.92  & 0.7888  & 27.78  & 0.7436  & 26.79  & 0.8068  & 31.18  & 0.9169  & 29.46  & 0.8313  \\
      & CASGCN \cite{J_PR_SR_CASGCN} & 32.60  & 0.9002  & 28.88  & 0.7890  & 27.70  & 0.7416  & 26.79  & 0.8086  & 31.18  & 0.9169  & 29.43  & 0.8313  \\
      & CSNLN \cite{SR:CSNLN} & 32.68  & 0.9004  & 28.95  & 0.7888  & 27.80  & 0.7439  & 27.22  & 0.8168  & 31.43  & 0.9201  & 29.62  & 0.8340  \\
      & SwinIR \cite{SwinIR} &\textcolor{blue}{32.72}  &\textcolor{red}{0.9021}  & 28.94  & 0.7914  & 27.83  & 0.7459  & 27.07  & 0.8164  & 31.67  &\textcolor{blue}{0.9226}  & 29.65  & 0.8357  \\
      & NLSN\cite{SR:NLSN} & 32.59  & 0.9000  & 28.87  & 0.7891  & 27.78  & 0.7444  & 26.96  & 0.8109  & 31.27  & 0.9184  & 29.49  & 0.8326  \\
      & DRLN \cite{SR:DRLN} & 32.63  & 0.9002  & 28.94  & 0.7900  & 27.83  & 0.7444  & 26.98  & 0.8119  & 31.54  & 0.9196  & 29.58  & 0.8332  \\
      & Restormer \cite{Restormer} & 32.12  & 0.8921  & 28.53  & 0.7757  & 27.70  & 0.7403  & 26.14  & 0.7805  & 30.95  & 0.9142  & 29.09  & 0.8205  \\
      & CRAN \cite{SR:CRAN} & 32.72  & 0.9012  &\textcolor{blue}{29.01}  & 0.7918  & 27.86  &\textcolor{blue}{0.7460}  & 27.13  & 0.8167  &\textcolor{blue}{31.75}  & 0.9219  &\textcolor{blue}{29.69}  & 0.8355  \\
      & TDPN \cite{SR:TDPN} & 32.69  & 0.9005  & 29.01  &\textcolor{red}{0.7943}  &\textcolor{red}{27.93}  & 0.7460  &\textcolor{blue}{27.24}  &\textcolor{blue}{0.8171}  & 31.58  & 0.9218  & 29.69  &\textcolor{blue}{0.8359}  \\
      & TFMAN(ours) &\textcolor{red}{32.75}  &\textcolor{blue}{0.9020}  &\textcolor{red}{29.03}  &\textcolor{blue}{0.7919}  &\textcolor{blue}{27.88}  &\textcolor{red}{0.7477}  &\textcolor{red}{27.27}  &\textcolor{red}{0.8205}  &\textcolor{red}{31.82}  &\textcolor{red}{0.9233}  &\textcolor{red}{29.75}  &\textcolor{red}{0.8371}  \\

      \hline
      \multirow{13}{*}{x8} 
      & Bicubic & 24.40  & 0.6580  & 23.10  & 0.5660  & 23.67  & 0.5480  & 20.74  & 0.5160  & 21.47  & 0.6500  & 22.68  & 0.5876  \\
      & SRCNN \cite{SR:SRCNN} & 25.33  & 0.6900  & 23.76  & 0.5910  & 24.13  & 0.5660  & 21.29  & 0.5440  & 22.46  & 0.6950  & 23.39  & 0.6172  \\
      & VDSR \cite{SR:VDSR} & 25.93  & 0.7240  & 24.26  & 0.6140  & 24.49  & 0.5830  & 21.70  & 0.5710  & 23.16  & 0.7250  & 23.91  & 0.6434  \\
      & LapSRN \cite{SR:LapSRN:Conf} & 26.15  & 0.7380  & 24.35  & 0.6200  & 24.54  & 0.5860  & 21.81  & 0.5810  & 23.39  & 0.7350  & 24.05  & 0.6520  \\
      & MemNet \cite{SR:MemNet} & 26.16  & 0.7414  & 24.38  & 0.6199  & 24.58  & 0.5842  & 21.89  & 0.5825  & 23.56  & 0.7387  & 24.11  & 0.6533  \\
      & EDSR \cite{SR:EDSR} & 26.96  & 0.7762  & 24.91  & 0.6420  & 24.81  & 0.5985  & 22.51  & 0.6221  & 24.69  & 0.7841  & 24.78  & 0.6846  \\
      & DBPN \cite{SR:DBPN} & 27.21  & 0.7840  & 25.13  & 0.6480  & 24.88  & 0.6010  & 22.73  & 0.6312  & 25.14  & 0.7987  & 25.02  & 0.6926  \\
      & RCAN \cite{SR:RCAN} & 27.31  & 0.7878  & 25.23  &\textcolor{blue}{0.6511}  & 24.98  &\textcolor{blue}{0.6058}  & 23.00  & 0.6452  & 25.24  & 0.8029  & 25.15  & 0.6986  \\
      & MS-LapSRN \cite{SR:LapSRN:Journ}& 26.34  & 0.7558  & 24.57  & 0.6273  & 24.65  & 0.5895  & 22.06  & 0.5963  & 23.90  & 0.7564  & 24.30  & 0.6651  \\
      & SAN \cite{SR:SAN} & 27.22  & 0.7829  & 25.14  & 0.6476  & 24.88  & 0.6011  & 22.70  & 0.6314  & 24.85  & 0.7906  & 24.96  & 0.6907  \\
      & CASGCN \cite{J_PR_SR_CASGCN} & 27.30  & 0.7849  & 25.23  & 0.6493  & 24.95  & 0.6034  & 23.01  & 0.6453  & 24.25  & 0.8034  & 24.95  & 0.6973  \\
      & DRLN \cite{SR:DRLN} &\textcolor{red}{27.36}  &\textcolor{red}{0.7882}  &\textcolor{red}{25.34}  &\textcolor{red}{0.6531}  &\textcolor{blue}{25.01}  & 0.6057  &\textcolor{blue}{23.06}  &\textcolor{blue}{0.6471}  &\textcolor{blue}{25.29}  &\textcolor{blue}{0.8041}  &\textcolor{blue}{25.21}  &\textcolor{red}{0.6996}  \\
      & TFMAN (ours) &\textcolor{blue}{27.34}  &\textcolor{blue}{0.7879}  &\textcolor{blue}{25.28}  & 0.6497  &\textcolor{red}{25.03}  &\textcolor{red}{0.6062}  &\textcolor{red}{23.13}  &\textcolor{red}{0.6482}  &\textcolor{red}{25.34}  &\textcolor{red}{0.8047}  &\textcolor{red}{25.23}  &\textcolor{blue}{0.6993}  \\
      
      \hline 
      \end{tabular} 
    
    \end{table*}

\begin{figure*}[!t]
    \centering
    \hspace{-0.04\linewidth}
    \subfloat[]{
      \includegraphics[width=0.33\linewidth]{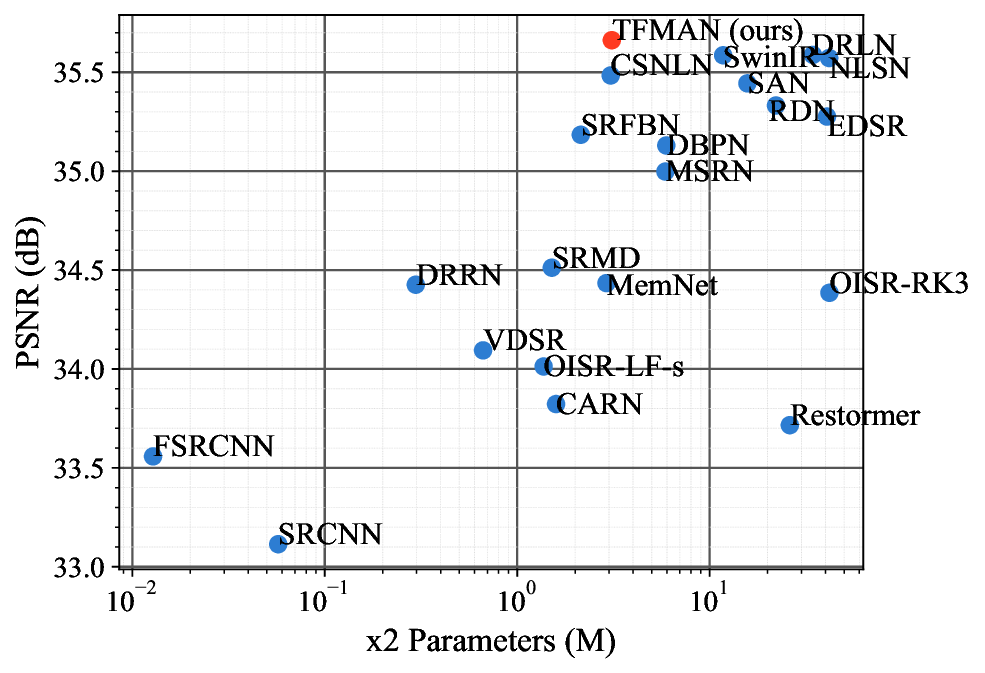}
      \label{fig_PSNR_vs_Params_x2}
    }\hspace{-0.03\linewidth}
    \subfloat[]{
      \includegraphics[width=0.33\linewidth]{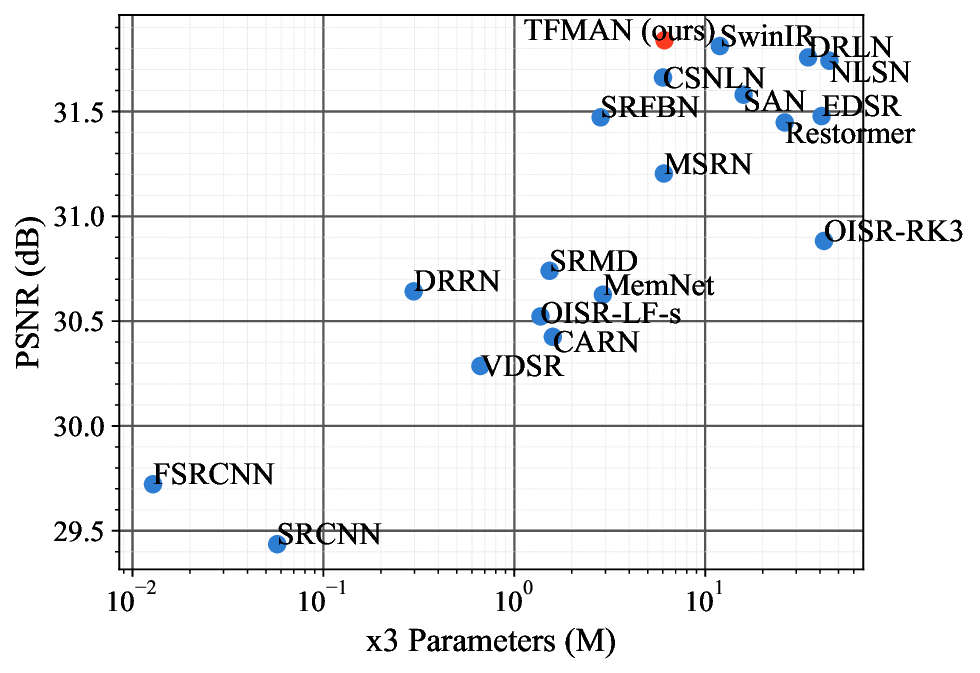}
      \label{fig_PSNR_vs_Params_x3}
    }\hspace{-0.03\linewidth}
    \subfloat[]{
      \includegraphics[width=0.33\linewidth]{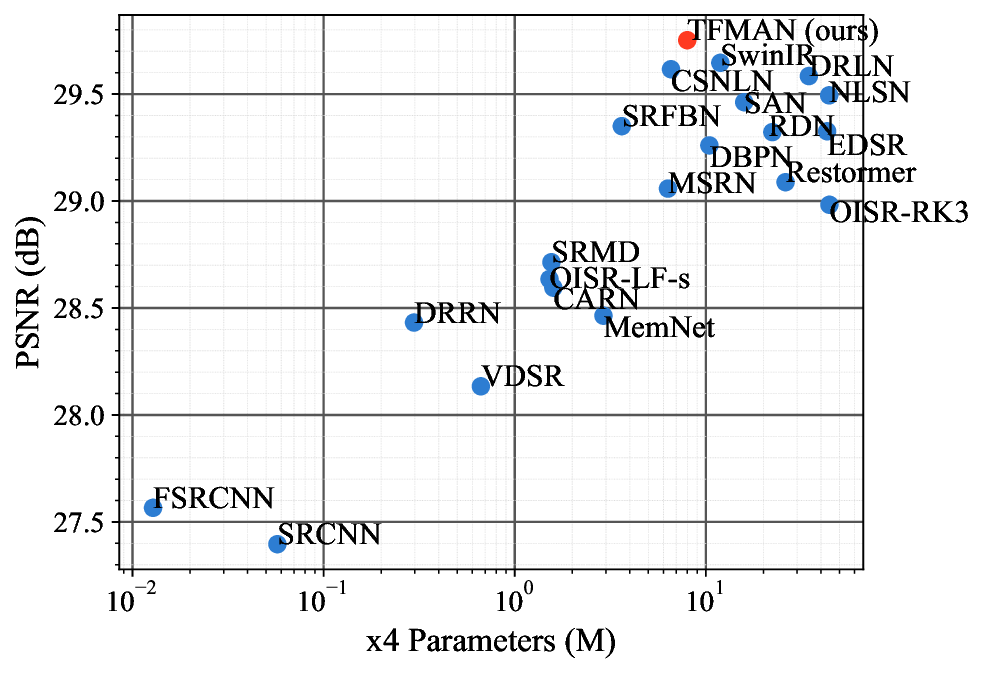}
      \label{fig_PSNR_vs_Params_x4}  
    }   
    \\%\vspace{-1mm}
    \hspace{-0.04\linewidth}
    \subfloat[]{
      \includegraphics[width=0.33\linewidth]{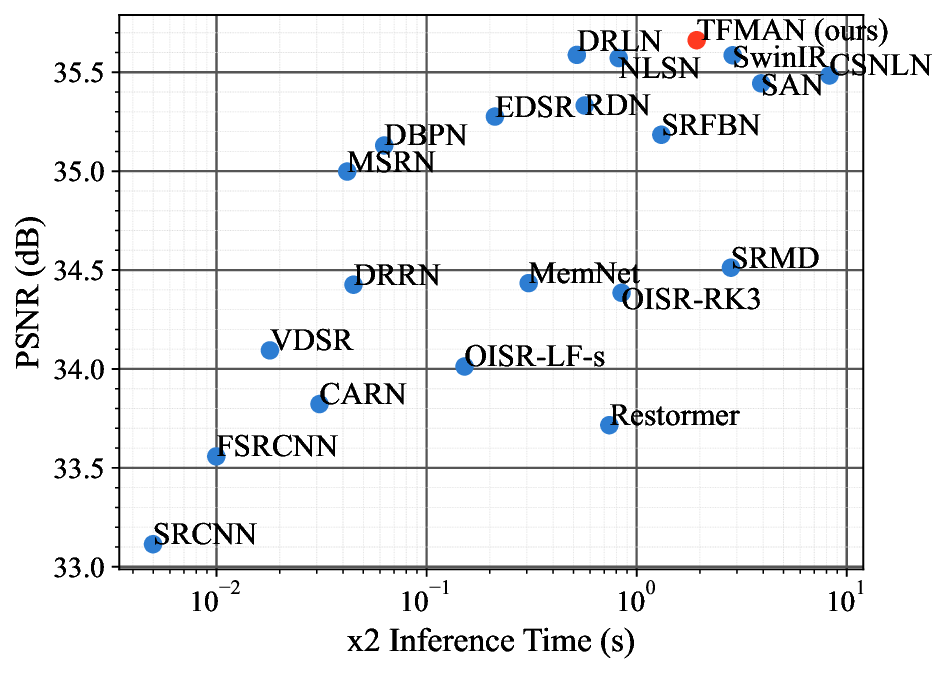}
      \label{fig_PSNR_vs_Time_x2}
    }\hspace{-0.03\linewidth}
    \subfloat[]{
      \includegraphics[width=0.33\linewidth]{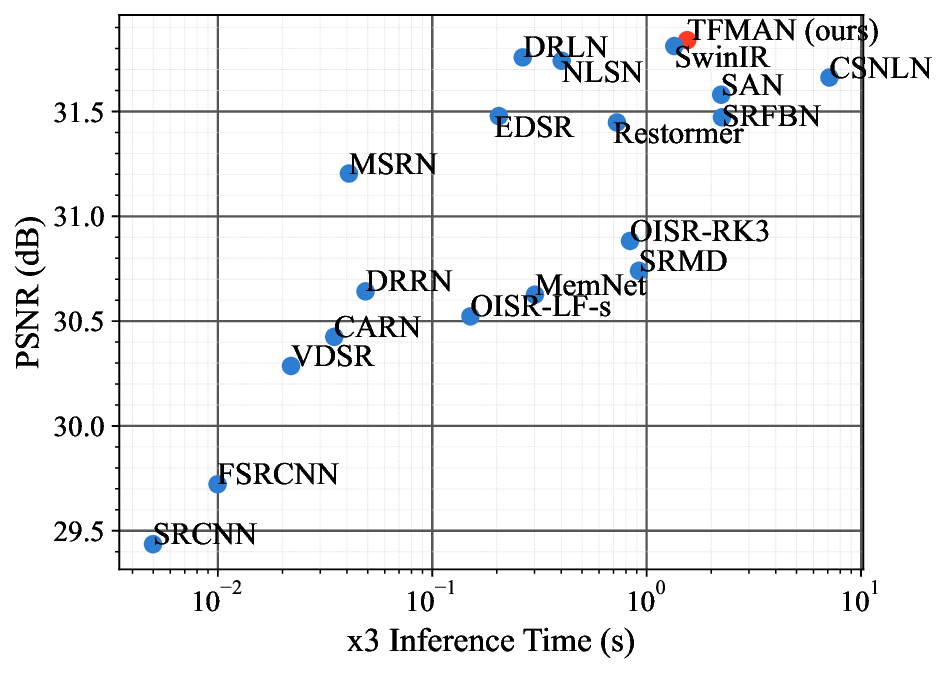}
      \label{fig_PSNR_vs_Time_x3}
    }\hspace{-0.03\linewidth}
    \subfloat[]{
      \includegraphics[width=0.33\linewidth]{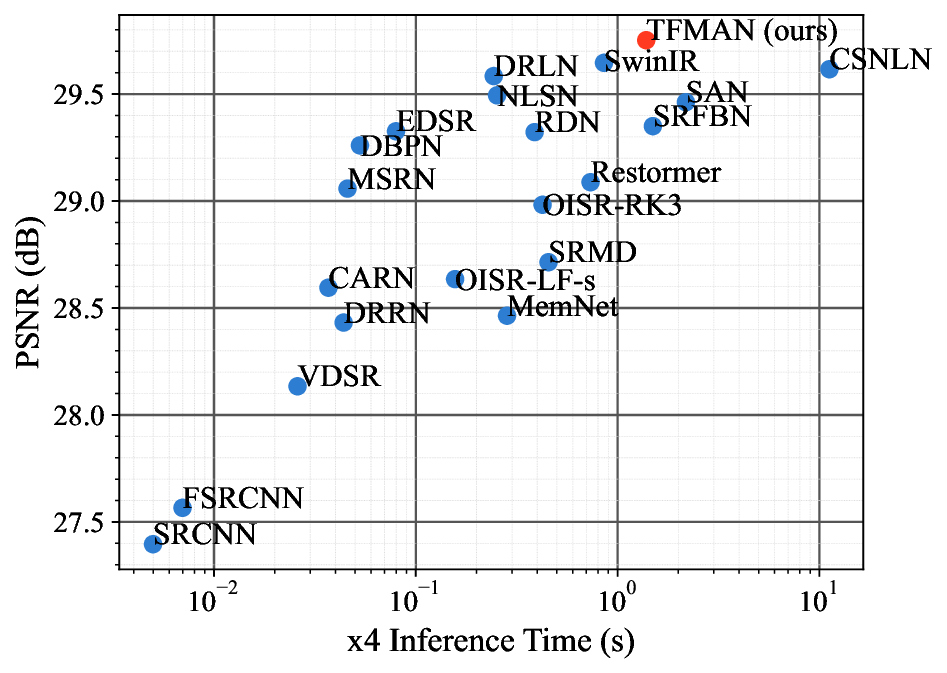}
      \label{fig_PSNR_vs_Time_x4}
    }\vspace{-2mm}
  
    \caption{
      The trade-offs between PSNR and parameters, 
      and between PSNR and inference time, for x2, x3 and x4.
      %     The upper part and the lower part respectively show the trade-offs between 
      % PSNR and parameters, and between PSNR and inference time
      % --- 
      The inference time is the total time spent inferring on Set5.
      The PSNR is the average of PSNR values 
      measured on the five benchmark datasets with BI degradation model.
      % In addition to the state-of-the-art methods that have exhibited in Table 
      % \ref{table_compare_BI}, 
      % some lightweight networks are also included for reference, including
      % MemNet \cite{SR:MemNet}.
      %         It should be noted that the inference speed of FSRCNN is slower than that 
      % of SRCNN. This is because FSRCNN contains far more convolution layers than 
      % SRCNN. In GPU, which is high parallel, the number of feedforward 
      % layers becomes the key to influence the inference speed.
    } 
    \label{fig_Model_complex_analyze}
  \end{figure*}

\subsubsection{Quantitative Results with BI}
As shown in Table \ref{table_compare_BI}, twenty state-of-the-art 
SR methods are compared quantitatively with TFMAN. 
They are 
SRCNN \cite{SR:SRCNN}, VDSR \cite{SR:VDSR}, LapSRN \cite{SR:LapSRN:Conf},
MemNet \cite{SR:MemNet}, EDSR \cite{SR:EDSR},  
DBPN \cite{SR:DBPN}, MS-LapSRN \cite{SR:LapSRN:Journ},
RCAN \cite{SR:RCAN}, OISR \cite{SR:OISR}, CSFM \cite{SR:CSFM}, 
RFANet \cite{SR:RFANet}, SAN \cite{SR:SAN}, CASGCN \cite{J_PR_SR_CASGCN}, 
CSNLN \cite{SR:CSNLN}, SwinIR \cite{SwinIR}, NLSN \cite{SR:NLSN}, 
DRLN \cite{SR:DRLN}, Restormer \cite{Restormer}, 
CRAN \cite{SR:CRAN} and TDPN \cite{SR:TDPN} for x2, x3, x4 and x8 SR.
Among them, because Restormer \cite{Restormer} has not been 
officially trained on SR tasks,
we add a nearest neighbor up-sampling to its head, and train it with
the training strategy of TFMAN for fair comparison.
We also made a joint comparison of model performance, 
parameters and inference time, 
which is shown in Figure \ref{fig_Model_complex_analyze}.
We also marked some methods that are not listed in Table, \ref{table_compare_BI}
including FSRCNN \cite{SR:FSRCNN}, CARN \cite{SR:CARN}, DRRN \cite{SR:DRRN}, 
SRMD \cite{SR:SRMD}, SRFBN \cite{SR:SRFBN} and MSRN \cite{SR:MSRN}.
Some methods are omitted in Figure \ref{fig_Model_complex_analyze}, 
since their paper does not expose code and related data,
and RCAN \cite{SR:RCAN} is due to location conflict with SAN \cite{SR:SAN}, etc.

In Table \ref{table_compare_BI}, PSNRs and SSIMs for scale factor 
over five benchmark datasets 
are reported. Most results are cited from their own papers. 
A few results that have not measured in their own papers are 
cited from the papers that completed the experiments. 
As reported in Table \ref{table_compare_BI}, our TFMAN is superior in most tests.
Of course, there are methods that surpass our TFMAN in 
individual indicators.
However, as shown in upper part of Figure \ref{fig_Model_complex_analyze}, 
the methods dominating in some cases have 
much more parameters than our method. 
For example, SwinIR \cite{SwinIR} has 11.7 million parameters and is 
around 4 times of our TFMAN that has only 3.1 million parameters for scale
factor of 2. 
As for CSNLN \cite{SR:CSNLN}, 
although it has achieved excellent results with a limited parameter count, 
as shown in lower part of Figure \ref{fig_Model_complex_analyze}, 
its inference speed is far behind the other methods, 
which takes 4-8 times as long as our method does. 
This is partly due to CSNLN's low parallelism of division strategy 
for region-level non-local.
From the above comparison, we can see that TFMAN not only has 
superior SR performance, but also takes fewer parameters, 
resulting in higher parameter utilization.

% review 2
\begin{figure*}[!t]
  \centering
  \renewcommand{\arraystretch}{0.6} %表示几倍行高
  \setlength\tabcolsep{2pt}
  \tiny

    % 一张图
    \begin{minipage}{0.247\textwidth} %大图
      \centering
      \begin{tabular}{c} 
        \includegraphics[width=\textwidth]{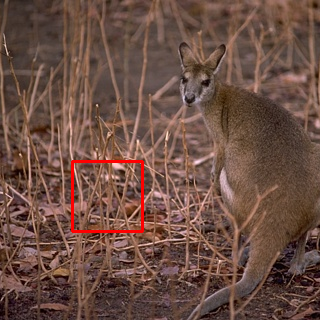}\\
        BSD100\\img\_090
      \end{tabular} 
    \end{minipage}\hspace{0mm}
    \begin{minipage}{0.69\textwidth} %小图
      \centering
      \begin{tabular}{*{6}{c}}  
        \includegraphics[width=0.15\textwidth]{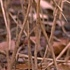} 
        &\includegraphics[width=0.15\textwidth]{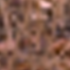}
        &\includegraphics[width=0.15\textwidth]{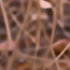}
        &\includegraphics[width=0.15\textwidth]{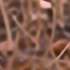}
        &\includegraphics[width=0.15\textwidth]{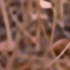}
        &\includegraphics[width=0.15\textwidth]{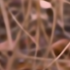} \\
        HR        &Bicubic      &RCAN \cite{SR:RCAN} &SRFBN \cite{SR:SRFBN} &OISR-RK3 \cite{SR:OISR} &SAN \cite{SR:SAN} \\
        PSNR/SSIM &24.43/0.5687 &25.37/0.6667        &25.26/0.6571          &25.31/0.6619            &25.39/0.6689\\
        \includegraphics[width=0.15\textwidth]{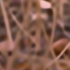}
        &\includegraphics[width=0.15\textwidth]{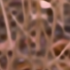}
        &\includegraphics[width=0.15\textwidth]{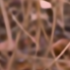}
        &\includegraphics[width=0.15\textwidth]{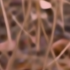}
        &\includegraphics[width=0.15\textwidth]{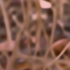}
        &\includegraphics[width=0.15\textwidth]{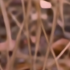} \\
        RFANet \cite{SR:RFANet} &CSNLN \cite{SR:CSNLN}    &DRLN \cite{SR:DRLN} &NLSN \cite{SR:NLSN} &SwinIR \cite{SwinIR} &TFMAN (Ours)  \\
        25.38/0.6659            &25.38/0.6654             &25.35/0.6646        &25.39/0.6664        &25.42/0.6701         &\textcolor{red}{25.49/0.6778} \\
      \end{tabular} 
    \end{minipage}

    % 一张图
    \begin{minipage}{0.247\textwidth} %大图
      \centering
      \begin{tabular}{c} 
        \includegraphics[width=\textwidth]{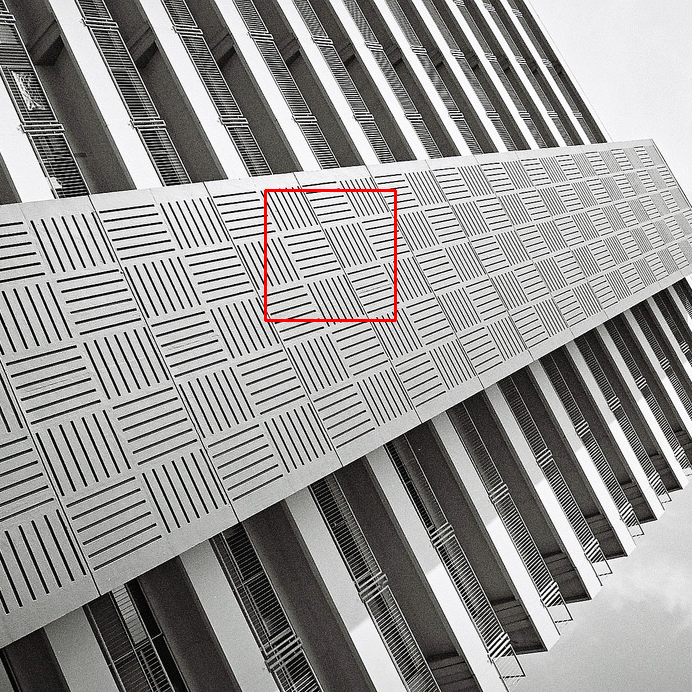}\\
        Urban100\\img\_092
      \end{tabular} 
    \end{minipage}\hspace{0mm}
    \begin{minipage}{0.69\textwidth} %小图
      \centering
      \begin{tabular}{*{6}{c}}  
        \includegraphics[width=0.15\textwidth]{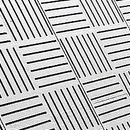} 
        &\includegraphics[width=0.15\textwidth]{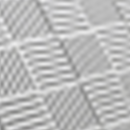}
        &\includegraphics[width=0.15\textwidth]{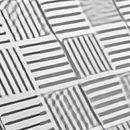}
        &\includegraphics[width=0.15\textwidth]{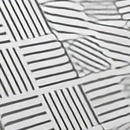}
        &\includegraphics[width=0.15\textwidth]{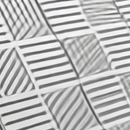}
        &\includegraphics[width=0.15\textwidth]{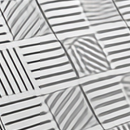} \\
        HR        &Bicubic      &RCAN \cite{SR:RCAN} &SRFBN \cite{SR:SRFBN} &OISR-RK3 \cite{SR:OISR} &SAN \cite{SR:SAN} \\
        PSNR/SSIM &16.61/0.4394 &19.67/0.6974        &19.55/0.6912          &19.18/0.6793            &19.53/0.6948\\
        \includegraphics[width=0.15\textwidth]{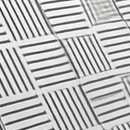}
        &\includegraphics[width=0.15\textwidth]{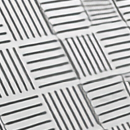}
        &\includegraphics[width=0.15\textwidth]{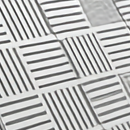}
        &\includegraphics[width=0.15\textwidth]{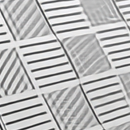}
        &\includegraphics[width=0.15\textwidth]{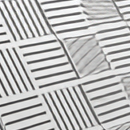}
        &\includegraphics[width=0.15\textwidth]{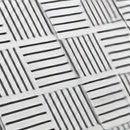} \\
        RFANet \cite{SR:RFANet} &CSNLN \cite{SR:CSNLN}    &DRLN \cite{SR:DRLN} &NLSN \cite{SR:NLSN} &SwinIR \cite{SwinIR} &TFMAN (Ours)  \\
        19.78/0.7026            &20.05/0.7120             &19.83/0.7028        &19.58/0.6964        &19.54/0.6910         &\textcolor{red}{20.17/0.7165} \\
      \end{tabular} 
    \end{minipage}

    % 一张图
    \begin{minipage}{0.247\textwidth} %大图
      \centering
      \begin{tabular}{c} 
        \includegraphics[width=\textwidth]{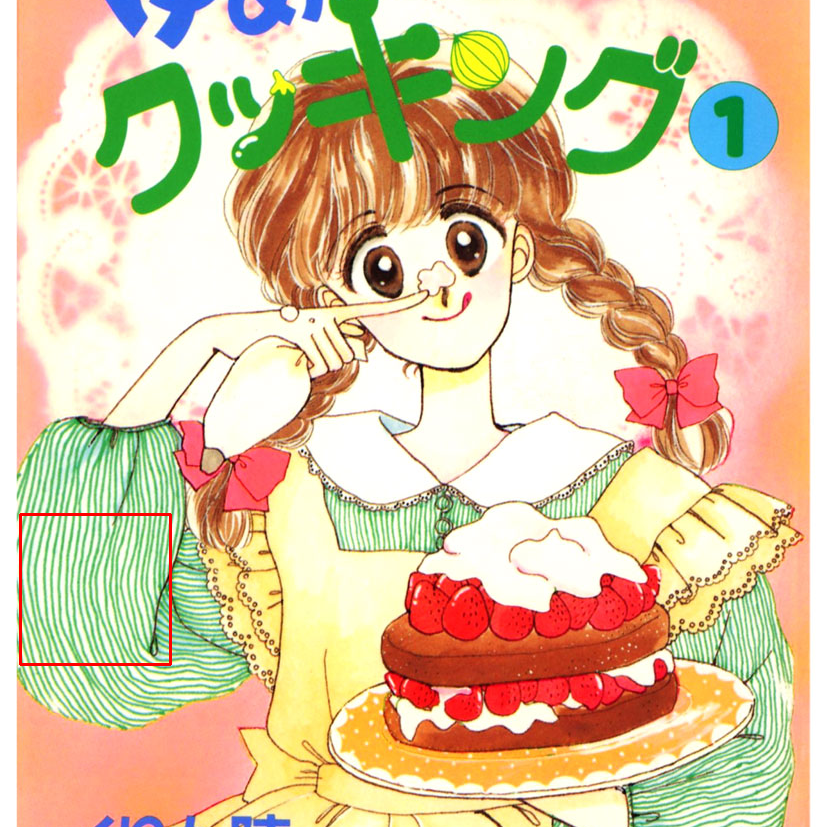}\\
        Manga109\\YumeiroCooking
      \end{tabular} 
    \end{minipage}\hspace{0mm}
    \begin{minipage}{0.69\textwidth} %小图
      \centering
      \begin{tabular}{*{6}{c}}  
        \includegraphics[width=0.15\textwidth]{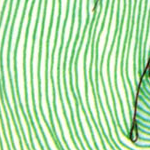} 
        &\includegraphics[width=0.15\textwidth]{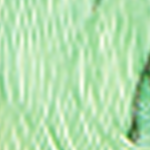}
        &\includegraphics[width=0.15\textwidth]{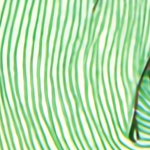}
        &\includegraphics[width=0.15\textwidth]{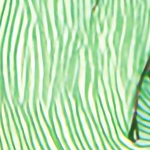}
        &\includegraphics[width=0.15\textwidth]{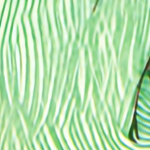}
        &\includegraphics[width=0.15\textwidth]{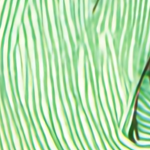} \\
        HR        &Bicubic      &RCAN \cite{SR:RCAN} &SRFBN \cite{SR:SRFBN} &OISR-RK3 \cite{SR:OISR} &SAN \cite{SR:SAN} \\
        PSNR/SSIM &24.76/0.7887 &29.85/0.9369        &28.68/0.9198          &28.69/0.9205            &29.15/0.9289\\
        \includegraphics[width=0.15\textwidth]{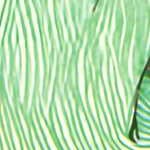}
        &\includegraphics[width=0.15\textwidth]{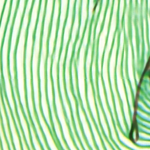}
        &\includegraphics[width=0.15\textwidth]{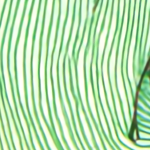}
        &\includegraphics[width=0.15\textwidth]{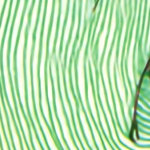}
        &\includegraphics[width=0.15\textwidth]{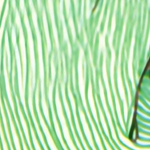}
        &\includegraphics[width=0.15\textwidth]{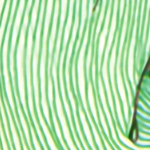} \\
        RFANet \cite{SR:RFANet} &CSNLN \cite{SR:CSNLN}    &DRLN \cite{SR:DRLN} &NLSN \cite{SR:NLSN} &SwinIR \cite{SwinIR} &TFMAN (Ours)  \\
        29.07/0.9251            &29.96/0.9394             &30.02/0.9392        &29.81/0.9377        &29.28/0.9292         &\textcolor{red}{30.42/0.9413} \\
      \end{tabular} 
    \end{minipage}

  \caption{
    Qualitative comparisons for x4 SR with BI degradation model.  
  } 
  \label{fig_qual_comp_x4}
\end{figure*}

% review 2
\begin{figure}[!t]
    \centering
    \renewcommand{\arraystretch}{0.4} %表示几倍行高
    \setlength\tabcolsep{1.45pt} % 列宽
    \tiny

    % 一张图
    \begin{minipage}{0.18\textwidth} %大图
      \begin{tabular}{c} 
        \includegraphics[width=\textwidth]{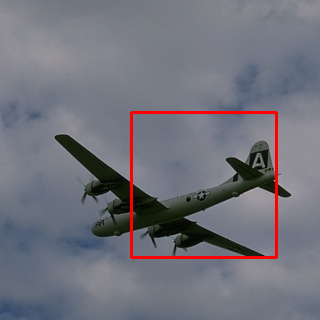}\\
        BSD100\\img\_067\\
      \end{tabular} 
    \end{minipage}\hspace{0.5mm}
    \begin{minipage}{0.31\textwidth} %小图
      \centering
      \begin{tabular}{cccc} 
        \includegraphics[width=0.232\textwidth]{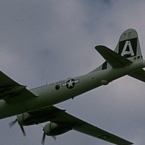} 
        &\includegraphics[width=0.232\textwidth]{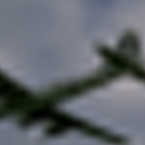} 
        &\includegraphics[width=0.232\textwidth]{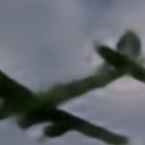} 
        &\includegraphics[width=0.232\textwidth]{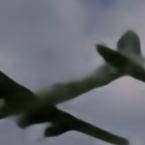}    \\
        HR        &Bicubic      &SRCNN \cite{SR:SRCNN}  &VDSR \cite{SR:VDSR} \\
        PSNR/SSIM &31.65/0.9324 &32.92/0.9451           &34.44/0.9575        \\ 
        \includegraphics[width=0.232\textwidth]{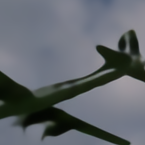} 
        &\includegraphics[width=0.232\textwidth]{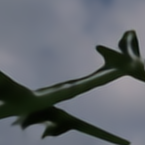} 
        &\includegraphics[width=0.232\textwidth]{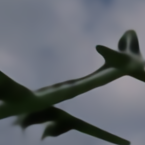} 
        &\includegraphics[width=0.232\textwidth]{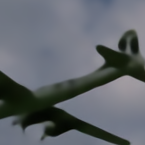}    \\
        DBPN \cite{SR:DBPN}   &RCAN \cite{SR:RCAN} &DRLN \cite{SR:DRLN}   &TFMAN \!\!(Ours)\\
        34.83/0.9628          &34.41/0.9618        &35.48/0.9642       &\textcolor{red}{35.88/0.9669}\\ 
      \end{tabular} 
    \end{minipage}

    % 一张图
    \begin{minipage}{0.18\textwidth} %大图
      \begin{tabular}{c} 
        \includegraphics[width=\textwidth]{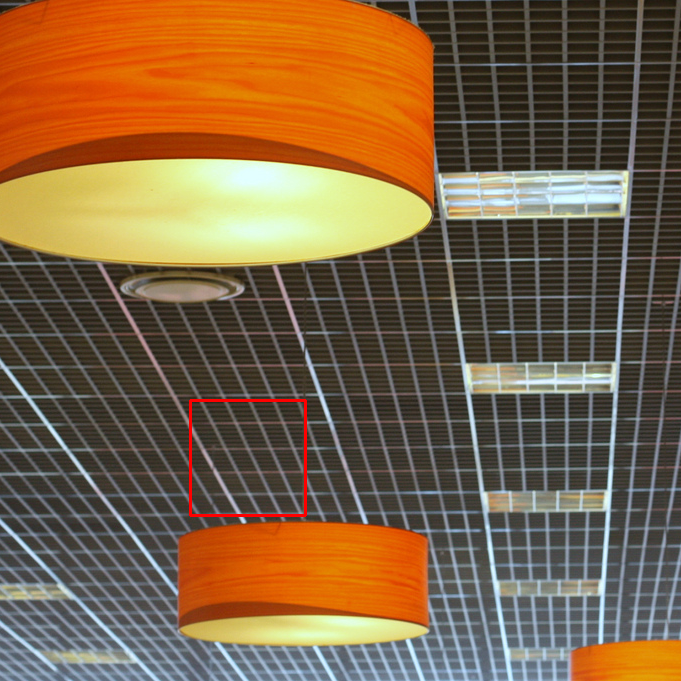}\\
        Urban100\\img\_044\\
      \end{tabular} 
    \end{minipage}\hspace{0.5mm}
    \begin{minipage}{0.31\textwidth} %小图
      \centering
      \begin{tabular}{cccc} 
        \includegraphics[width=0.232\textwidth]{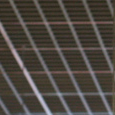} 
        &\includegraphics[width=0.232\textwidth]{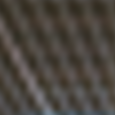} 
        &\includegraphics[width=0.232\textwidth]{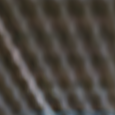} 
        &\includegraphics[width=0.232\textwidth]{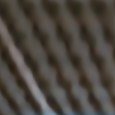}    \\
        HR        &Bicubic      &SRCNN \cite{SR:SRCNN}  &VDSR \cite{SR:VDSR} \\
        PSNR/SSIM &22.82/0.4934 &23.58/0.5405           &23.73/0.5383        \\ 
        \includegraphics[width=0.232\textwidth]{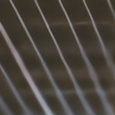} 
        &\includegraphics[width=0.232\textwidth]{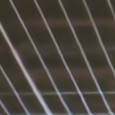} 
        &\includegraphics[width=0.232\textwidth]{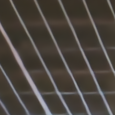} 
        &\includegraphics[width=0.232\textwidth]{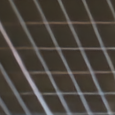}    \\
        DBPN \cite{SR:DBPN}   &RCAN \cite{SR:RCAN} &DRLN \cite{SR:DRLN}   &TFMAN \!\!(Ours)\\
        25.59/0.6756          &27.35/0.7412        &27.31/0.7394       &\textcolor{red}{27.88/0.7589}\\ 
      \end{tabular} 
    \end{minipage}

    \caption{
      Qualitative comparisons for x8 SR with BI degradation model.  
    } 
    \label{fig_qual_comp_x8}
  \end{figure}

\subsubsection{Qualitative Results with BI}
We qualitatively compare TFMAN with several representative state-of-the-art 
SR methods for x4 and x8 SR.

Figure \ref{fig_qual_comp_x4} shows the comparison for x4 SR. 
The first image is a low resolution photo taken from nature scenarios. 
Therefore, it is difficult to reconstruct the textures when scale factor is 
large. 
It can be seen that the stem stripes recovered by our method are the clearest.
Urban100 is characterized by an abundance of lines and 
geometries at different scales.
Therefore, methods exploiting image self-similarity
should play a better performance on this dataset, 
such as CSNLN \cite{SR:CSNLN}, NLSN \cite{SR:NLSN} and SwinIR \cite{SwinIR}.
However, it can be clearly seen that TFMAN is the only method recovering 
the vertical lines of the upper right tile in the second image.
This demonstrates the effectiveness to explicitly explore and  
preserve the training features for CNN-based methods.
Manga109 is characterized by irregular hand-drawn lines.
In the third image, only RCAN \cite{SR:RCAN}, CSNLN \cite{SR:CSNLN},
DRLN \cite{SR:DRLN} and TFMAN can clearly recover the lines on the girl's sleeve. 
However, careful comparison shows that the trend of the lines reconstructed 
by TFMAN is closer to the original.
The comparison for more challenging x8 SR 
is shown in Figure \ref{fig_qual_comp_x8}.
In the first image, due to the loss of too much information, 
methods is too difficult to recover details for the aircraft
and even generates bad artifacts.
Our method is the only one that does not generate white strip artifacts
among the four newly proposed methods.
As for the second picture, 
only our method can clearly recover the ceiling stripes.
The above visual comparisons strongly demonstrate 
the effectiveness of our TFMAN.

% Review 3
\begin{table*}[!t]
    \renewcommand{\arraystretch}{0.6} % 表示几倍的行高
    \caption{Quantitative comparison with BD and DN degradation model for 
    x3. 
    The red and the blue index respectively denote the best and 
  the second best results.} 
    \label{table_compare_BD_DN}
    \centering
    \tiny
     
    \begin{tabular}{|*{15}{@{}c@{}|}} %表格格式在此定义
      \hline
      \multirow{2}{*}{Degrad.}&\multirow{2}{*}{Method}&
      \multicolumn{2}{c|}{Set5}&\multicolumn{2}{c|}{Set14}&\multicolumn{2}{c|}{BSD100}&
      \multicolumn{2}{c|}{Urban100}&\multicolumn{2}{c|}{Manga109}&\multicolumn{2}{c|}{Average}\\
      \cline{3-14}
      &&PSNR&SSIM&PSNR&SSIM&PSNR&SSIM&PSNR&SSIM&PSNR&SSIM&PSNR&SSIM\\
  
      \hline
      \multirow{12}{*}{BD} 
  
      & Bicubic & 28.78  & 0.8308  & 26.38  & 0.7271  & 26.33  & 0.6918  & 23.52  & 0.6862  & 25.46  & 0.8149  & 26.09  & 0.7502  \\
      & SPMSR \cite{SR:SPMSR} & 32.21  & 0.9001  & 28.89  & 0.8105  & 28.13  & 0.7740  & 25.84  & 0.7856  & 29.64  & 0.9003  & 28.94  & 0.8341  \\
      & IRCNN \cite{SR:IRCNN} & 33.38  & 0.9182  & 29.63  & 0.8281  & 28.65  & 0.7922  & 26.77  & 0.8154  & 31.15  & 0.9245  & 29.92  & 0.8557  \\
      & RDN \cite{SR:RDN:Conf} & 34.58  & 0.9280  & 30.53  & 0.8447  & 29.23  & 0.8079  & 28.46  & 0.8582  & 33.97  & 0.9465  & 31.35  & 0.8771  \\
      & SRFBN \cite{SR:SRFBN}& 34.66  & 0.9283  & 30.48  & 0.8439  & 29.21  & 0.8069  & 28.48  & 0.8581  & 34.07  & 0.9466  & 31.38  & 0.8768  \\
      & RCAN \cite{SR:RCAN}& 34.70  & 0.9288  & 30.63  & 0.8462  & 29.32  & 0.8093  & 28.81  & 0.8647  & 34.38  & 0.9483  & 31.57  & 0.8795  \\
      & SAN \cite{SR:SAN}& 34.75  & 0.9290  & 30.68  & 0.8466  & 29.33  & 0.8101  & 28.83  & 0.8646  & 34.46  & 0.9487  & 31.61  & 0.8798  \\
      & RFANet \cite{SR:RFANet}& 34.77  & 0.9292  & 30.68  & 0.8473  & 29.34  & 0.8104  & 28.89  & 0.8661  & 34.49  & 0.9492  & 31.63  & 0.8804  \\
      & CASGCN \cite{J_PR_SR_CASGCN} & 34.62  & 0.9283  & 30.60  & 0.8458  & 29.30  &\textcolor{red}{0.8196}  & 28.68  & 0.8611  & 34.27  & 0.9476  & 31.48  & 0.8805  \\
      & Restormer \cite{Restormer} & 34.64  & 0.9282  & 30.60  & 0.8447  & 29.27  & 0.8083  & 28.59  & 0.8574  & 34.28  & 0.9476  & 31.47  & 0.8773  \\
      & DRLN \cite{SR:DRLN} &\textcolor{blue}{34.81}  &\textcolor{blue}{0.9297}  &\textcolor{blue}{30.81}  &\textcolor{red}{0.8487}  &\textcolor{blue}{29.40}  &\textcolor{blue}{0.8121}  &\textcolor{blue}{29.11}  &\textcolor{blue}{0.8697}  &\textcolor{blue}{34.84}  &\textcolor{blue}{0.9506}  &\textcolor{blue}{31.79}  &\textcolor{blue}{0.8822}  \\
      & TFMAN (ours) &\textcolor{red}{34.84}  &\textcolor{red}{0.9299}  &\textcolor{red}{30.81}  &\textcolor{blue}{0.8486}  &\textcolor{red}{29.40}  & 0.8120  &\textcolor{red}{29.27}  &\textcolor{red}{0.8726}  &\textcolor{red}{34.88}  &\textcolor{red}{0.9512}  &\textcolor{red}{31.84}  &\textcolor{red}{0.8828}  \\
      \hline

      \hline
      \multirow{9}{*}{DN} 
      & Bicubic & 24.14  & 0.5445  & 23.14  & 0.4828  & 22.94  & 0.4461  & 21.63  & 0.4701  & 23.08  & 0.5448  & 22.99  & 0.4977  \\
      & SRCNN \cite{SR:SRCNN} & 27.16  & 0.7672  & 25.49  & 0.6580  & 25.11  & 0.6151  & 23.32  & 0.6500  & 25.78  & 0.7889  & 25.37  & 0.6958  \\
      & VDSR \cite{SR:VDSR} & 27.72  & 0.7872  & 25.92  & 0.6786  & 25.52  & 0.6345  & 23.83  & 0.6797  & 26.41  & 0.8130  & 25.88  & 0.7186  \\
      & IRCNN \cite{SR:IRCNN} & 26.18  & 0.7430  & 24.68  & 0.6300  & 24.52  & 0.5850  & 22.63  & 0.6205  & 24.74  & 0.7701  & 24.55  & 0.6697  \\
      & SRMD \cite{SR:SRMD} & 27.74  & 0.8026  & 26.13  & 0.6974  & 25.64  & 0.6495  & 24.28  & 0.7092  & 26.72  & 0.8424  & 26.10  & 0.7402  \\
      & RDN \cite{SR:RDN:Conf} & 28.46  &\textcolor{blue}{0.8151}  & 26.60  & 0.7101  & 25.93  & 0.6573  & 24.92  & 0.7362  & 28.00  & 0.8590  & 26.78  & 0.7555  \\
      & SRFBN \cite{SR:SRFBN} & 28.53  &\textcolor{red}{0.8182}  & 26.60  &\textcolor{red}{0.7144}  & 25.95  &\textcolor{red}{0.6625}  & 24.99  &\textcolor{blue}{0.7424}  & 28.02  &\textcolor{blue}{0.8618}  & 26.82  &\textcolor{blue}{0.7599}  \\
      & Restormer \cite{Restormer} &\textcolor{red}{28.66}  & 0.8144  &\textcolor{blue}{26.69}  & 0.7090  &\textcolor{blue}{25.98}  & 0.6599  &\textcolor{blue}{25.16}  & 0.7395  &\textcolor{blue}{28.16}  & 0.8602  &\textcolor{blue}{26.93}  & 0.7566  \\
      & TFMAN (ours) &\textcolor{blue}{28.65}  & 0.8142  &\textcolor{red}{26.72}  &\textcolor{blue}{0.7113}  &\textcolor{red}{25.99}  &\textcolor{blue}{0.6616}  &\textcolor{red}{25.46}  &\textcolor{red}{0.7521}  &\textcolor{red}{28.39}  &\textcolor{red}{0.8661}  &\textcolor{red}{27.04}  &\textcolor{red}{0.7611}  \\
      \hline

      \end{tabular} 
    
    \end{table*}

\subsubsection{Quantitative Results with BD and DN}
Table \ref{table_compare_BD_DN} reports the quantitative results of models 
with BD and DN degradation. 

For BD degradation model, DRLN \cite{SR:DRLN} has achieved  
high performance on each dataset. However, our method is better than
DRLN in almost all benchmarks.
Especially for Urban100, TFMAN exceeds the second 
place by 0.16 in PSNR, which is a huge improvement.
For the DN degradation model, 
among the models, Restormer \cite{Restormer} is the newly proposed 
state-of-the-art method for image restoration.
Although it did not stand out in the above SR experiments, 
it performed well on DN tasks combining denoising and SR.
However, TFMAN achieves better. Although TFMAN does not achieve the best 
results on several SSIM indicators, our method far exceeds the 
second place in almost all PSNR indicators.
In particular, Urban100 and Manga109 respectively exceeded the second 
place by 0.30 and 0.23.
The above comparisons show that TFMAN not only is superior in recovering 
lost details coming from image reduction, but also has excellent 
performance in image deblurring and denoising.

% Review 3
\begin{figure}[!t]
    \centering
    \renewcommand{\arraystretch}{0.4} %表示几倍行高
    \setlength\tabcolsep{1.45pt} % 列宽
    \tiny

    % 一张图
    \begin{minipage}{0.18\textwidth} %大图
      % \centering
      \begin{tabular}{c} 
        \includegraphics[width=\textwidth]{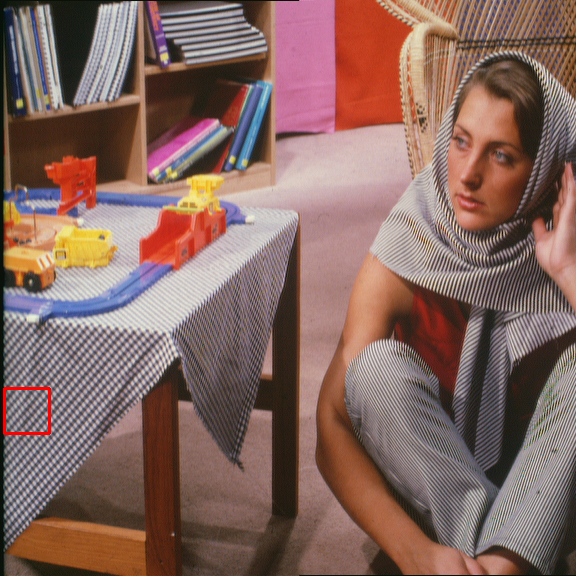}\\
        Set14\\barbara\\
      \end{tabular} 
    \end{minipage}\hspace{0.25mm}
    \begin{minipage}{0.31\textwidth} %小图
      \centering
      \begin{tabular}{cccc} 
        \includegraphics[width=0.232\textwidth]{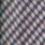} 
        &\includegraphics[width=0.232\textwidth]{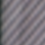} 
        &\includegraphics[width=0.232\textwidth]{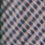} 
        &\includegraphics[width=0.232\textwidth]{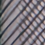}    \\
        HR        &Bicubic      &RCAN \cite{SR:RCAN} &SRFBN \cite{SR:SRFBN} \\
        PSNR/SSIM &25.60/0.7157 &27/42/0.8179        &27/15/0.8071        \\ 
        \includegraphics[width=0.232\textwidth]{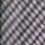} 
        &\includegraphics[width=0.232\textwidth]{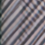} 
        &\includegraphics[width=0.232\textwidth]{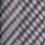} 
        &\includegraphics[width=0.232\textwidth]{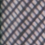}    \\
        RFANet \cite{SR:RFANet} &DRLN \cite{SR:DRLN} &Restormer \cite{Restormer} &TFMAN \!\!(Ours)\\
        27.50/0.8202            &27.39/0.8168        &27.38/0.8105               &\textcolor{red}{27.53/0.8230}\\ 
      \end{tabular} 
    \end{minipage}

    % 一张图
    \begin{minipage}{0.18\textwidth} %大图
      % \centering
      \begin{tabular}{c} 
        \includegraphics[width=\textwidth]{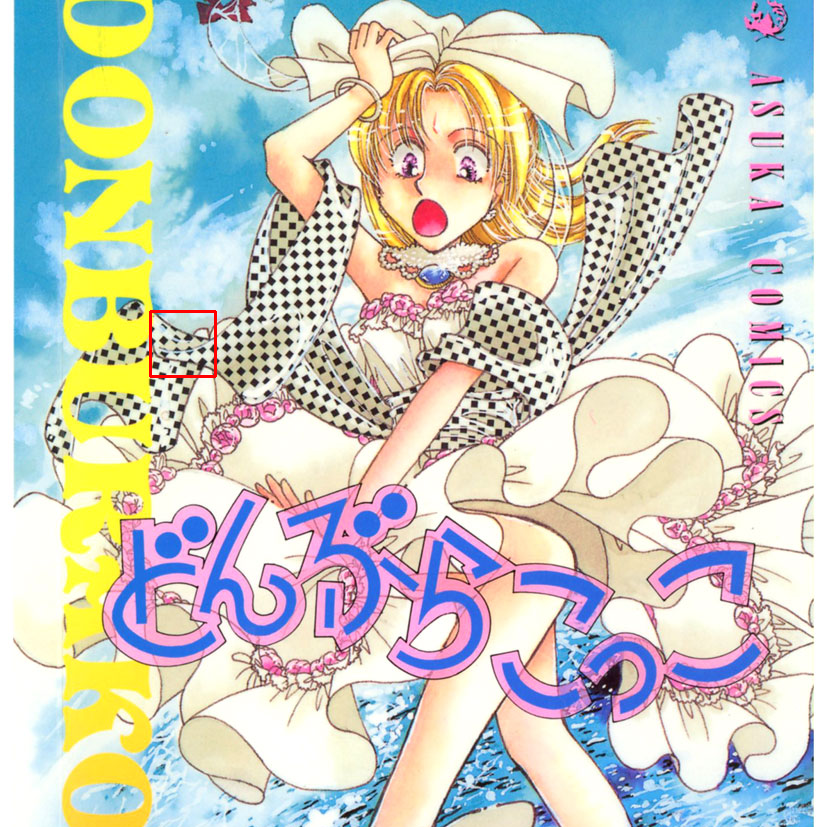}\\
        Manga109\\Donburakokko\\
      \end{tabular} 
    \end{minipage}\hspace{0.25mm}
    \begin{minipage}{0.31\textwidth} %小图
      \centering
      \begin{tabular}{cccc} 
        \includegraphics[width=0.232\textwidth]{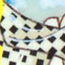} 
        &\includegraphics[width=0.232\textwidth]{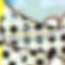} 
        &\includegraphics[width=0.232\textwidth]{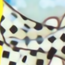} 
        &\includegraphics[width=0.232\textwidth]{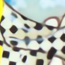}    \\
        HR        &Bicubic      &RCAN \cite{SR:RCAN} &SRFBN \cite{SR:SRFBN} \\
        PSNR/SSIM &24.15/0.7911 &31.57/0.9549        &31.36/0.9531        \\ 
        \includegraphics[width=0.232\textwidth]{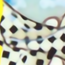} 
        &\includegraphics[width=0.232\textwidth]{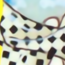} 
        &\includegraphics[width=0.232\textwidth]{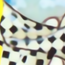} 
        &\includegraphics[width=0.232\textwidth]{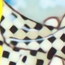}    \\
        RFANet \cite{SR:RFANet} &DRLN \cite{SR:DRLN} &Restormer \cite{Restormer} &TFMAN \!\!(Ours)\\
        31.65/0.9555            &31.99/0.9575        &31.49/0.9544               &\textcolor{red}{32.12/0.9581}\\ 
      \end{tabular} 
    \end{minipage}

    \caption{
      Qualitative comparisons for x3 SR with BD degradation model.  
    } 
    \label{fig_qual_comp_BD}
  \end{figure}

% Review 3
\begin{figure}[!t]
    \centering
    \renewcommand{\arraystretch}{0.4} %表示几倍行高
    \setlength\tabcolsep{1.45pt} % 列宽
    \tiny

    % 一张图
    \begin{minipage}{0.18\textwidth} %大图
      \begin{tabular}{c} 
        \includegraphics[width=\textwidth]{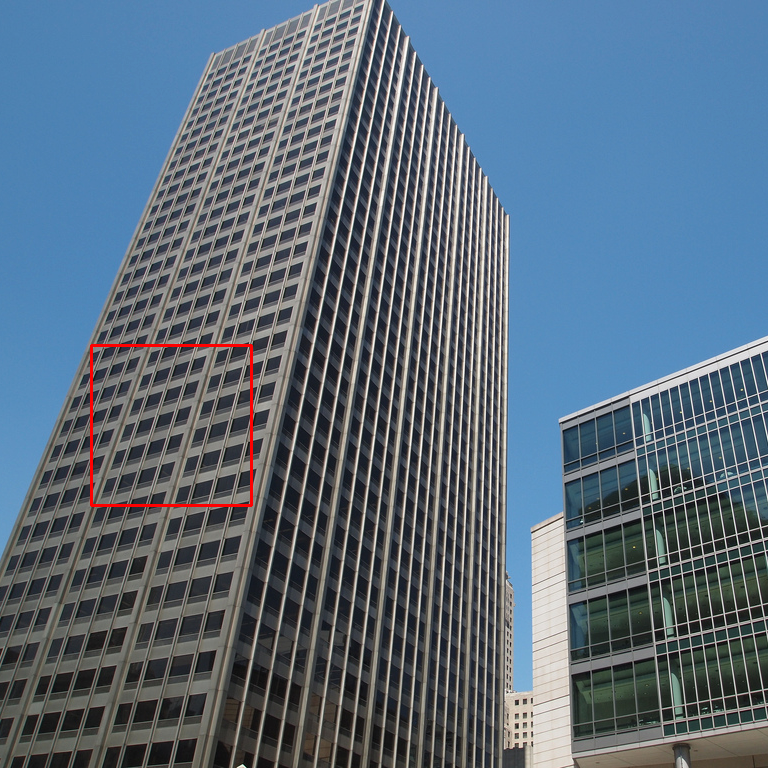}\\
        Urban100\\img\_096\\
      \end{tabular} 
    \end{minipage}\hspace{0.5mm}
    \begin{minipage}{0.31\textwidth} %小图
      \centering
      \begin{tabular}{cccc} 
        \includegraphics[width=0.232\textwidth]{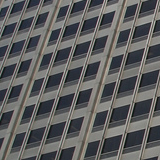} 
        &\includegraphics[width=0.232\textwidth]{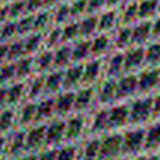} 
        &\includegraphics[width=0.232\textwidth]{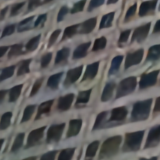} 
        &\includegraphics[width=0.232\textwidth]{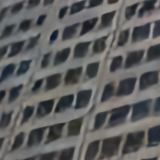}    \\
        HR        &Bicubic      &VDSR \cite{SR:VDSR}  &SRMD \cite{SR:SRMD} \\
        PSNR/SSIM &20.77/0.4464 &23.38/0.7806         &23.92/0.7984        \\ 
        \includegraphics[width=0.232\textwidth]{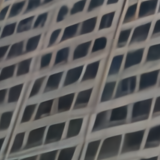} 
        &\includegraphics[width=0.232\textwidth]{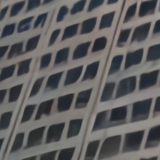} 
        &\includegraphics[width=0.232\textwidth]{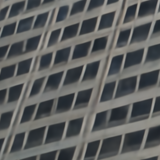} 
        &\includegraphics[width=0.232\textwidth]{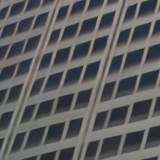}    \\
        RDN \cite{SR:RDN:Conf}   &SRFBN \cite{SR:SRFBN} &Restormer \cite{Restormer}   &TFMAN \!\!(Ours)\\
        25.26/0.8401             &24.81/0.8336          &25.07/0.8395              &\textcolor{red}{25.92/0.8629}\\ 
      \end{tabular} 
    \end{minipage}

    % 一张图
    \begin{minipage}{0.18\textwidth} %大图
      \begin{tabular}{c} 
        \includegraphics[width=\textwidth]{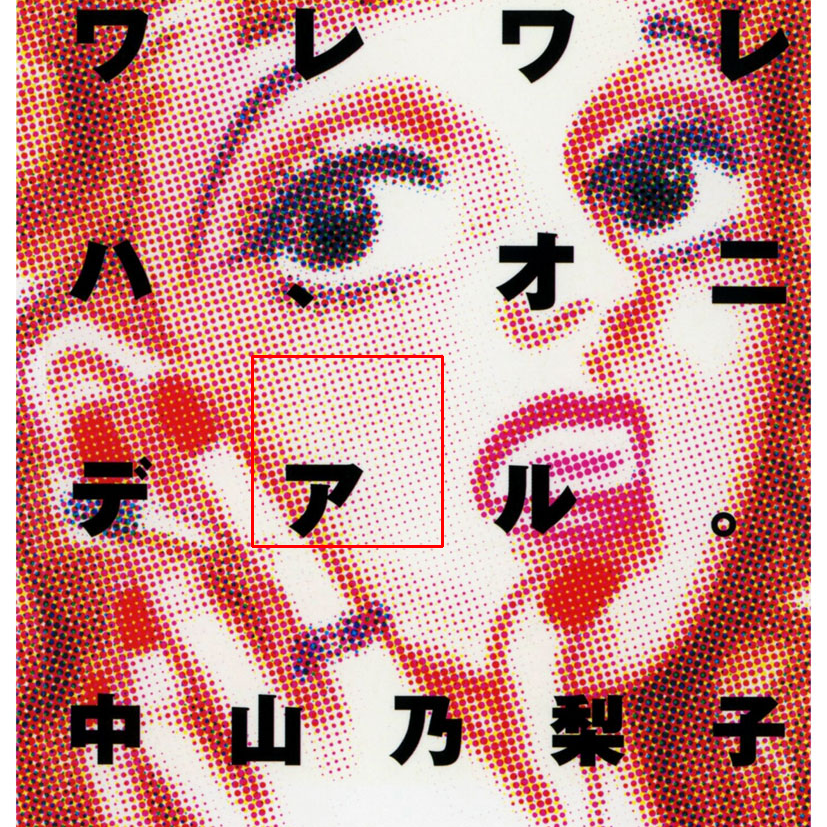}\\
        Manga109\\WarewareHaOniDearu\\
      \end{tabular} 
    \end{minipage}\hspace{0.5mm}
    \begin{minipage}{0.31\textwidth} %小图
      \centering
      \begin{tabular}{cccc} 
        \includegraphics[width=0.232\textwidth]{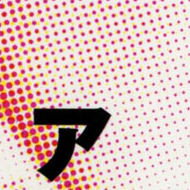} 
        &\includegraphics[width=0.232\textwidth]{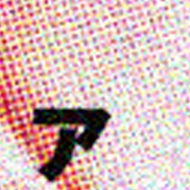} 
        &\includegraphics[width=0.232\textwidth]{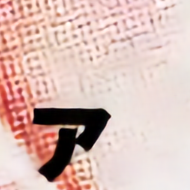} 
        &\includegraphics[width=0.232\textwidth]{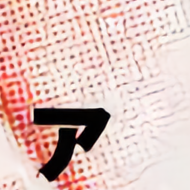}    \\
        HR        &Bicubic      &VDSR \cite{SR:VDSR}  &SRMD \cite{SR:SRMD} \\
        PSNR/SSIM &21.71/0.6657 &21.31/0.6961         &22.04/0.7538        \\ 
        \includegraphics[width=0.232\textwidth]{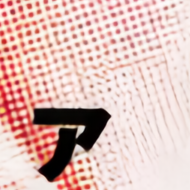} 
        &\includegraphics[width=0.232\textwidth]{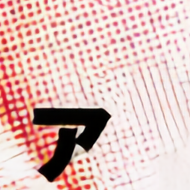} 
        &\includegraphics[width=0.232\textwidth]{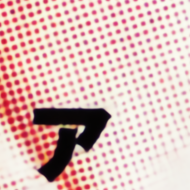} 
        &\includegraphics[width=0.232\textwidth]{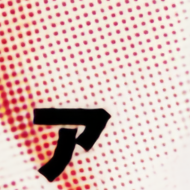}    \\
        RDN \cite{SR:RDN:Conf}   &SRFBN \cite{SR:SRFBN} &Restormer \cite{Restormer}   &TFMAN \!\!(Ours)\\
        22.80/0.7903             &22.41/0.7857          &24.15/0.8381                 &\textcolor{red}{24.39/0.8540}\\ 
      \end{tabular} 
    \end{minipage}

    \caption{
      Qualitative comparisons for x3 SR with DN degradation model.  
    }
     
    \label{fig_qual_comp_DN}
  
  \end{figure}

\subsubsection{Qualitative Results with BD and DN}
Figure \ref{fig_qual_comp_BD} and \ref{fig_qual_comp_DN} respectively show
the visual results with BD and DN degradation models.

In Figure \ref{fig_qual_comp_BD}, it can be seen that 
TFMAN produces sharper outlines of the 
tablecloth and the clothes of the girl than other methods do.
Moreover, through careful comparison, it can be found that our 
method even recovers the degradation of the HR images that may 
be defocus blur or be introduced from image compression algorithm.
In Figure \ref{fig_qual_comp_DN}, for the first image, our method 
not only effectively removes the noise, but also accurately restores 
the geometric shapes of the windows.
For the second image, on the basis of removing the noise, TFMAN also 
preserves the red dots used to make up the cartoon, which is likely 
to be removed as noise.
For example, SRCNN \cite{SR:SRCNN} and VDSR \cite{SR:VDSR} mistakenly remove a 
large region of red dots, resulting in their PSNR scores even lower than Bicubic.
The above experiments demonstrate that TFMAN not 
only effectively eliminates the image degradation, 
but also preserves the texture of the original image.

\subsection{Model Analysis}
In this subsection, we first explore the effectiveness of each module in TFMAN. 
Afterwards, we delve into the analysis of the trade-off between the model's
depth and the channel number of middle features.
Without loss of generality, these experiments are 
conducted with scale factor of 2 and BI degradation model.

% Review 3
\begin{table*}[!t]
    \renewcommand{\arraystretch}{0.6} %表示几倍行高
    \caption{
      Ablation study on modules of TFMAN. PSNR results and total inference 
      time on Set5 are reported after 15000 epochs of training.  
    } 
    \label{table_ablation_comparison}
    \centering
    \tiny
  
    \begin{tabular}{|*{10}{@{}c@{}|}} %表格格式在此定义
      \hline 
      Module&Base&R1&R2&R3&Original&R5&R6&R7&R8\\
      
      \hline 
      TFM	&		&	\Checkmark	&		&		&	\Checkmark	&		&	\Checkmark	&	\Checkmark	&	S	\\
      SRNL	&		&		&	\Checkmark	&		&	\Checkmark	&	\Checkmark	&		&	\Checkmark	&	\Checkmark	\\
      CA	&		&		&		&	\Checkmark	&	\Checkmark	&	\Checkmark	&	\Checkmark	&		&	\Checkmark	\\
      \hline
      Inference Time (s) & 1.079&1.535&1.575&1.084&1.940&1.577&1.537&1.934&1.856 \\ 
      Parameters (M)	&	3.7327 	&	3.2146 	&	3.6182 	&	3.7349 	&	3.1022 	&	3.6204 	&	3.2168 	&	3.1001 	&	4.8176 	\\
      PSNR (dB)	&	38.04&38.07&38.17&38.05&38.36&38.20&38.10&38.25&38.26 \\
      \hline 
      \end{tabular} 
  \end{table*}

\subsubsection{Ablation Study}
As reported in Table \ref{table_ablation_comparison}, the ablation study is 
conducted to explore the influence of TFM, SRNL and CA on our model.
The base model is built by replacing TFM, SRNL and CA of the original model with 
$6\times 6$ deconvolution, $3\times 3$ convolution, and identity connection, respectively.
In order to demonstrate that TFMAN can achieve a higher model capacity through
channel-independent feature matching, we introduce an additional training, 
labeled R8, whose TFM module simultaneously matches features across all 
channels at a position. We set its feature number $N = 512$ and feature 
channel number $R = 8$. Before matching features, its channel 
number $R$ is first mapped to 128 through a $1\times 1$ convolution. 

By comparing Base, R1 and R2, after the convolutional operations are respectively
replaced by TFM and SRNL, the reduction of model parameters and the enhanced SR 
performance verify the effectiveness of these two modules.
In fact, through this experiment, we can clearly calculate that the parameters
of matching feature sets in TFM is only
\begin{equation}
  nNRs^2K^2=12\times 32\times 4\times 2^2\times 3^2=55296.
\end{equation}
Of course, because the computational requirements of the two modules are 
larger than those of convolutions, the inference time is also relatively 
increased. When comparing R1 and R3, the close PSNR indicates that the 
solely added CA has no significant effect to model performance.
However, when TFM and CA are combined, as seen in the comparison between  
Base and R6, or R2 and Original, the performance is greatly improved. 
This demonstrates the complementary nature of TFM and CA.
Furthermore, the comparison between R8 and the original model indicates  
that channel-independent matching performs better.

\begin{figure}[!t]
    \centering
    \includegraphics[width=0.45\linewidth]{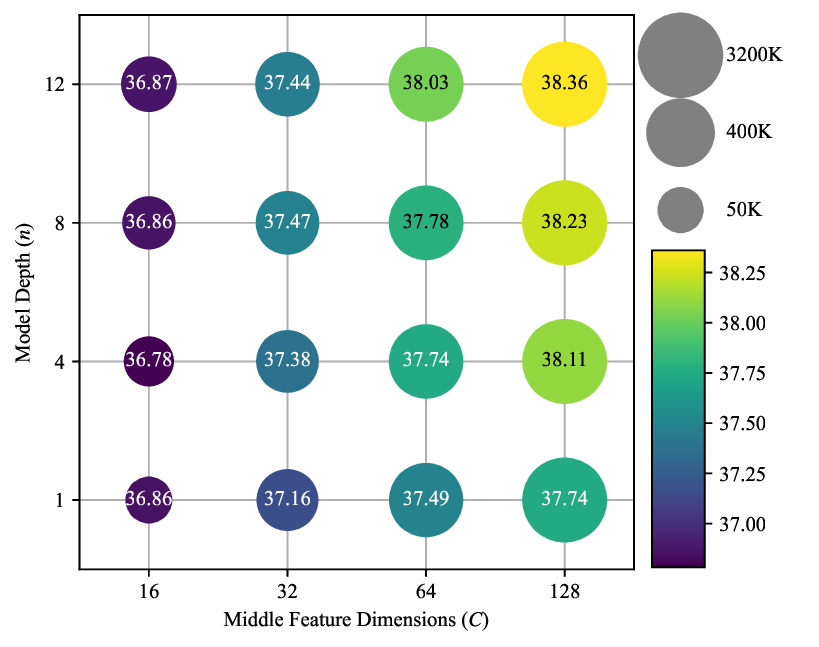} 
    \caption{
      The study for impact of recurrence depth and middle feature number to TFMAN.
      PSNR results on Set5 are recorded after 15000 epochs.
          The size of the circle and the intensity of its color respectively 
      represent the number of parameters and PSNR value.
    } 
    \label{fig_model_size_analysis}
  \end{figure} 

\subsubsection{Model Size Study}
We further investigate the impact of the trade-off between the depth of model
recurrence and the middle feature channel number on TFMAN.
Figure \ref{fig_model_size_analysis} presents the experimental 
results for different combinations of depth $n$ and channel dimension $C$.
Vertically, the number of model parameters is similar for a fixed feature dimension. 
As the depth increases, the model's performance will 
eventually reach a peak value constrained by its parameters,
as shown in the first and second columns.
Horizontally, as the feature dimension increases, 
the model's performance consistently improves for a fixed depth.
However, it should be noted that the channel dimension grows exponentially, 
so it will soon hit the performance bottleneck of hardware.
Additionally, with increasing feature dimensions, one can observe that 
the optimal result among a series of models with a smaller feature dimension 
approaches or even surpasses the suboptimal result among a series of
models with a larger feature dimension.
Therefore, it can be predicted that the performance of our model with 12
recurrence depth and 128 feature dimensions should be comparable to that
of the model with 4 recurrence depth and 256 feature dimensions,
which was not included in the experiment due to equipment limitation.

\subsection{Module Explorations}

\begin{figure}[!t]
    \centering
    \includegraphics[width=0.45\linewidth]{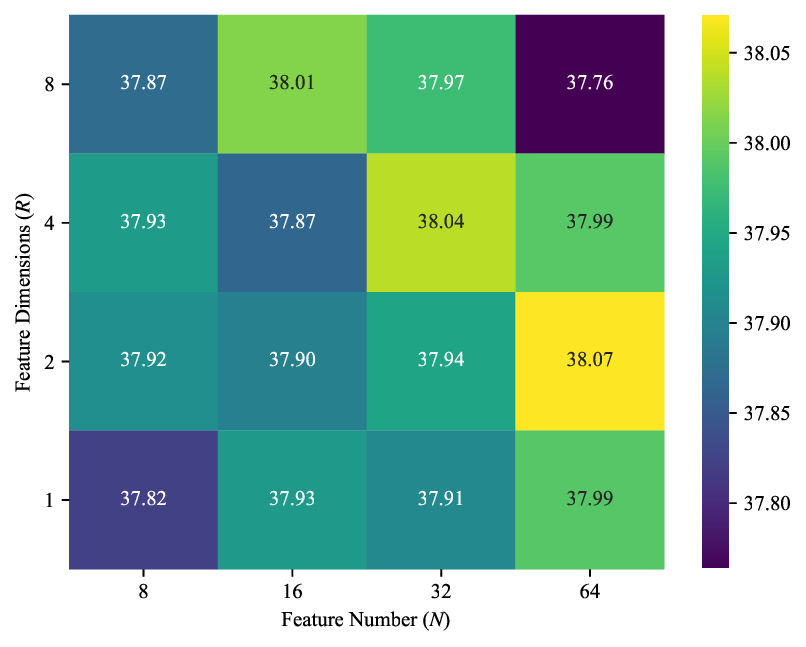} 
    \caption{
      Exploration of TFM volume.
      PSNR results on Set5 are recorded after 15000 epochs, which is
      represented by the intensity of the grids' color.
    } 
    \label{fig_TFM_size_analysis}
  \end{figure}

% Review 2
\begin{figure*}[!t]
    \centering
    \renewcommand{\arraystretch}{0.6} %表示几倍行高
    \setlength\tabcolsep{1.5pt} % 列宽
    \scriptsize

    % 一张图
    \begin{minipage}{0.206\textwidth} %大图
      \centering
      \begin{tabular}{c} 
        \includegraphics[width=\textwidth]{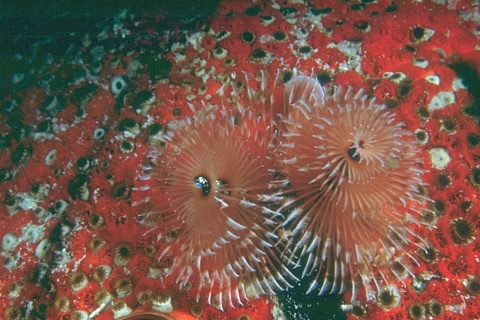}\\
        BSD100 img\_012\\
      \end{tabular} 
    \end{minipage}%\hspace{-3mm}
    \begin{minipage}{0.750\textwidth} %小图
      \centering
      \begin{tabular}{*{8}{c}}  
        \includegraphics[width=0.111\textwidth]{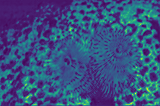} 
        &\includegraphics[width=0.111\textwidth]{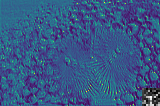} 
        &\includegraphics[width=0.111\textwidth]{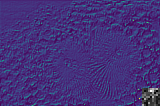}
        &\includegraphics[width=0.111\textwidth]{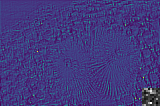}
        &\includegraphics[width=0.111\textwidth]{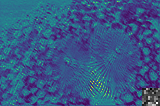}
        &\includegraphics[width=0.111\textwidth]{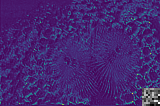}
        &\includegraphics[width=0.111\textwidth]{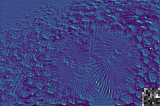}
        &\includegraphics[width=0.111\textwidth]{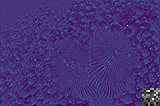}  \\
        Input C1&Act F1&Act F2&Act F3&Act F4&Act F5&Act F6&Act F7 \\
        
        \includegraphics[width=0.111\textwidth]{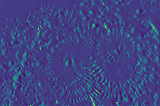} 
        &\includegraphics[width=0.111\textwidth]{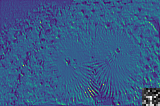} 
        &\includegraphics[width=0.111\textwidth]{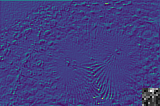}
        &\includegraphics[width=0.111\textwidth]{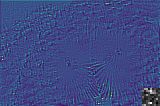}
        &\includegraphics[width=0.111\textwidth]{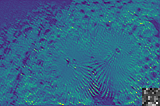}
        &\includegraphics[width=0.111\textwidth]{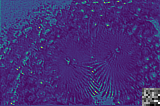}
        &\includegraphics[width=0.111\textwidth]{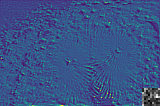}
        &\includegraphics[width=0.111\textwidth]{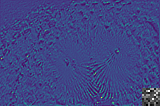}  \\
        Input C2&Act F1&Act F2&Act F3&Act F4&Act F5&Act F6&Act F7 \\
      \end{tabular} 
    \end{minipage}
  
    \vspace{0.4mm}
    \centering
    % 一张图
    \begin{minipage}{0.206\textwidth} %大图
      \centering
      \begin{tabular}{c} 
        \includegraphics[width=\textwidth]{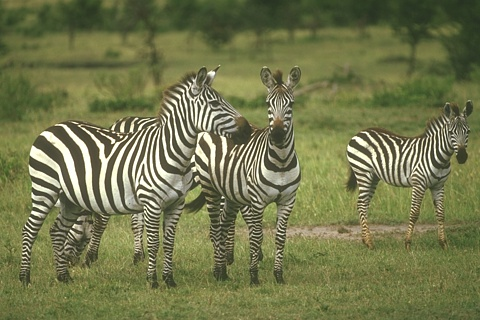}\\
        BSD100 img\_052\\
      \end{tabular} 
    \end{minipage}%\hspace{-3mm}
    \begin{minipage}{0.750\textwidth} %小图
      \centering
      \begin{tabular}{*{8}{c}}  
        \includegraphics[width=0.111\textwidth]{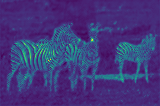} 
        &\includegraphics[width=0.111\textwidth]{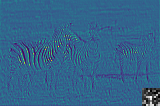}
        &\includegraphics[width=0.111\textwidth]{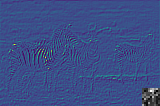}
        &\includegraphics[width=0.111\textwidth]{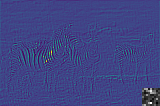}
        &\includegraphics[width=0.111\textwidth]{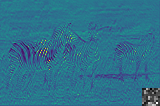}
        &\includegraphics[width=0.111\textwidth]{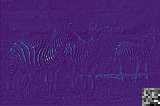}
        &\includegraphics[width=0.111\textwidth]{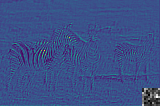}
        &\includegraphics[width=0.111\textwidth]{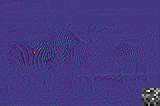}  \\
        Input C1&Act F1&Act F2&Act F3&Act F4&Act F5&Act F6&Act F7 \\
        
        \includegraphics[width=0.111\textwidth]{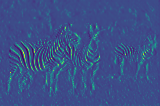} 
        &\includegraphics[width=0.111\textwidth]{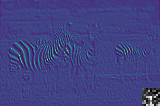}
        &\includegraphics[width=0.111\textwidth]{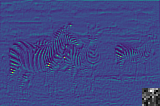}
        &\includegraphics[width=0.111\textwidth]{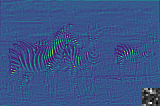}
        &\includegraphics[width=0.111\textwidth]{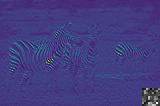}
        &\includegraphics[width=0.111\textwidth]{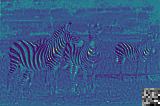}
        &\includegraphics[width=0.111\textwidth]{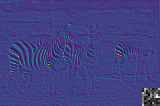}
        &\includegraphics[width=0.111\textwidth]{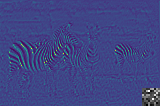}\\ 
        Input C2&Act F1&Act F2&Act F3&Act F4&Act F5&Act F6&Act F7 \\
      \end{tabular} 
    \end{minipage}

    \caption{Visualization of channel matchings in TFM.
    The visualization is implemented on model with scale factor of 3 and BI degradation.
    Input C$x$ represents the $x$-th channel of the input feature map. 
    Act F$x$ represents the matching of the $x$-th feature to the corresponding 
    channel in the same line, and the feature is shown in the lower right corner.
    } 
    \label{fig_TFM_visualization}
  \end{figure*}

\subsubsection{Exploration of TFM}
We first study the effect of feature set volume on the performance of TFMAN.
The experiment is implemented with scale factor of 2 and BI degradation model.
To expedite training, 
we reduce the middle feature channel number $C$ from 128 to 64.
Figure \ref{fig_TFM_size_analysis} shows the experimental results of
TFMANs applying TFM with different combinations of feature number $N$ 
and feature channel number $R$. 
When examining the diagonal direction, it can be concluded that 
the feature sets with same volume yield similar effects on the model,
and those with more features are slightly dominant.
Furthermore, it is evident that the feature set volume doesn't 
follow a linear "bigger is better" relationship. Instead, the PSNR
peaks when $N\times R=128$, and works best when $N=64, R=2$.
Considering that the computational load of TFM is dominated by 
the feature number $N$, we finally adopt TFM with $N = 32$ and $R = 4$ to 
strike a balance between performance and computation.

We further visualize the TFM module to intuitively understand its feature matching.
Figure \ref{fig_TFM_visualization} shows the matching results of 
the first seven features with the first two channels 
of the input feature map in the second recurrence.
Limited by the definition of feature matching operation, 
the features within TFM feature sets might not exhibit 
straightforward geometric properties, as is observed in 
methods based on sparse coding.
In fact, as per its definition, the features learned 
by TFM can be conceptualized as linear combinations 
of various elementary feature decompositions.
Additionally, 
since each matching feature tends to emphasize distinct texture,
and different channels are dedicated to diverse features,
the module can generate a multitude of enhancements, 
totaling $C \times N = 128 \times 32$, 
which greatly improves the representation ability of the model.

\begin{figure*}
    \centering
    \subfloat[]{
      \includegraphics[width=0.31\linewidth]{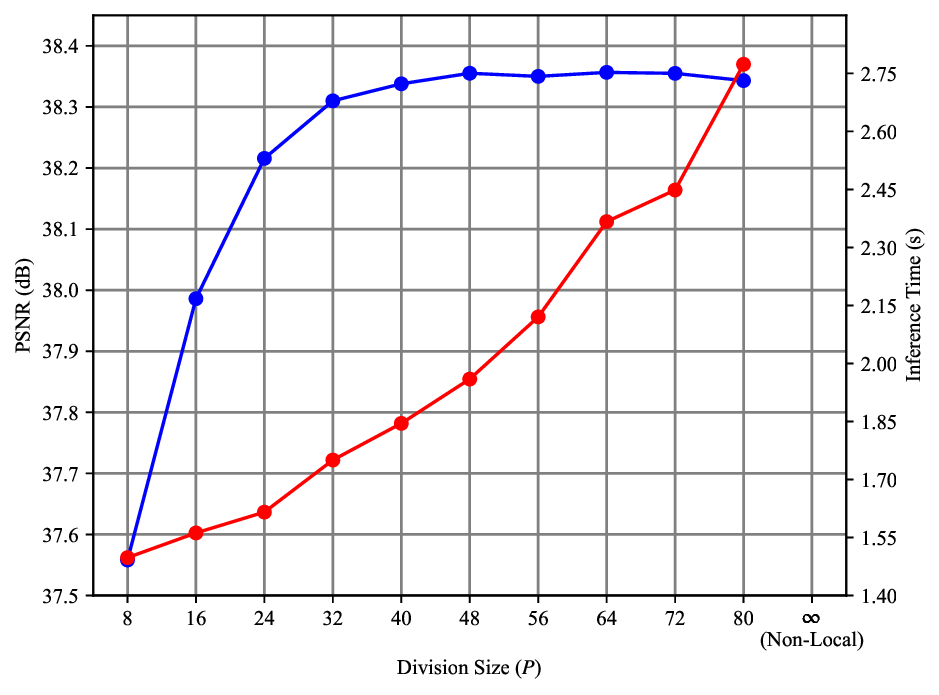}
      \label{fig_SRNL_patch_size_comp_x2}
    }\hspace{-0.01\linewidth}
    \subfloat[]{
      \includegraphics[width=0.31\linewidth]{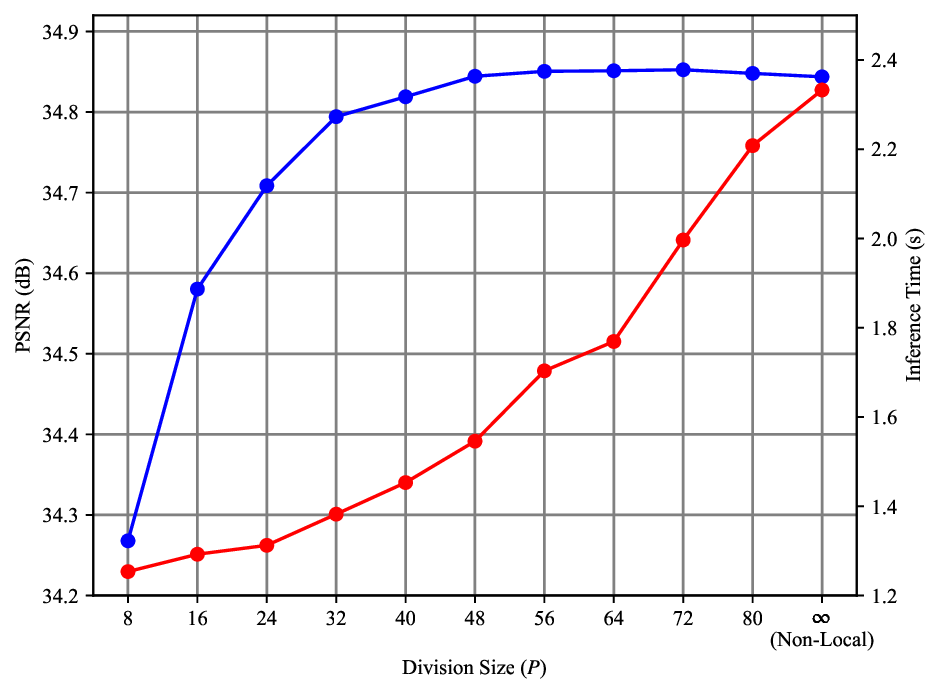}
      \label{fig_SRNL_patch_size_comp_x3}
    }\hspace{-0.01\linewidth}
    \subfloat[]{
      \includegraphics[width=0.31\linewidth]{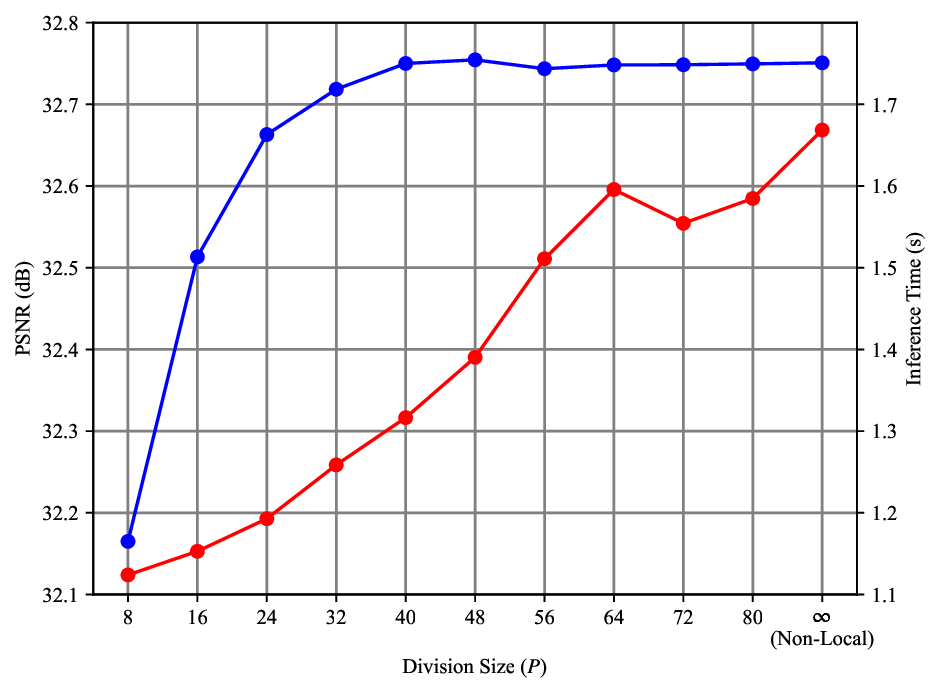}
      \label{fig_SRNL_patch_size_comp_x4}
    }
  
    \caption{
      Exploration for division size of SRNL. Sub-figure (a), (b) and (c) 
      are respectively the records for x2, x3 and x4. 
          The red line and the blue line respectively indicate the change 
      of inference time and PSNRs of TFMAN measuring on Set5.
                        } 
    \label{fig_SRNL_patch_size_comp}
  \end{figure*}

% Review 3
\begin{figure*}[!t]
    \centering
    \renewcommand{\arraystretch}{0.5} %表示几倍行高
    \setlength\tabcolsep{0.8pt} % 列宽
    \tiny

    % 一张图
    \begin{minipage}{0.2\textwidth} %大图
      \centering
      \begin{tabular}{c} 
        \includegraphics[width=\textwidth]{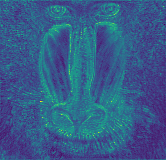}\\
        Input Feature Map\\
      \end{tabular} 
    \end{minipage}\hspace{3mm}
    \begin{minipage}{0.2\textwidth} %小图
      \centering
      \begin{tabular}{cccc}%{|ccc|c|}  
        % \hline
        \includegraphics[width=0.23\textwidth]{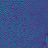} 
        &\includegraphics[width=0.23\textwidth]{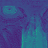}
        &\includegraphics[width=0.23\textwidth]{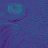} 
        &\includegraphics[width=0.23\textwidth]{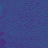}  \\
        
        \includegraphics[width=0.23\textwidth]{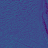} 
        &\includegraphics[width=0.23\textwidth]{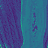}
        &\includegraphics[width=0.23\textwidth]{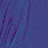}
        &\includegraphics[width=0.23\textwidth]{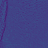}  \\
        
        \includegraphics[width=0.23\textwidth]{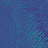} 
        &\includegraphics[width=0.23\textwidth]{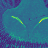}
        &\includegraphics[width=0.23\textwidth]{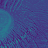}
        &\includegraphics[width=0.23\textwidth]{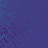}  \\ %\hline
        
        \includegraphics[width=0.23\textwidth]{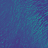} 
        &\includegraphics[width=0.23\textwidth]{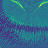}
        &\includegraphics[width=0.23\textwidth]{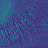}
        &\includegraphics[width=0.23\textwidth]{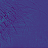}  \\ %\hline
        \multicolumn{4}{c}{Blocks After Non-Local} \\
      \end{tabular} 
    \end{minipage}\hspace{3mm}
    \begin{minipage}{0.2\textwidth} %大图
      \centering
      \begin{tabular}{c} 
        \includegraphics[width=\textwidth]{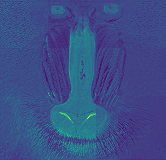}\\
        Merged Blocks\\
      \end{tabular} 
    \end{minipage}\hspace{3mm}
    \begin{minipage}{0.2\textwidth} %大图
      \centering
      \begin{tabular}{c} 
        \includegraphics[width=\textwidth]{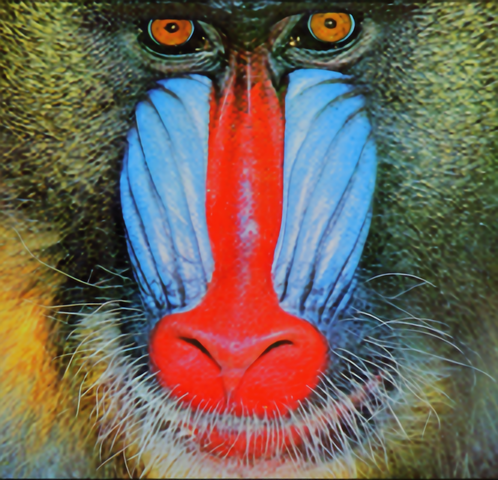}\\
        Final SR Result\\
      \end{tabular} 
    \end{minipage}%\hspace{-3mm}
  
    \caption{
      Visualization of SRNL. 
      The first image is the channel average of the input 
      feature map of SRNL.
      The second image is the results of divided blocks after performing non-local.
      The third image is the result after merging these blocks.
      The fourth image is the final x3 SR result of TFMAN.
    } 
    \label{fig_SRNL_visualization}
  \end{figure*}

\subsubsection{Exploration of SRNL}
\label{SRNL_Exploration}
Figure \ref{fig_SRNL_patch_size_comp} depicts the change in PSNR 
and inference time of TFMAN when transitioning from utilizing 
SRNL with an $8\times 8$ division size to directly employing  
non-local for scale factors of 2, 3 and 4 on the Set5 dataset.
One can find that when the division size is around 48, 
TFMAN achieves its best performance, accompanied by a 
reduction in inference time compared to larger division sizes.
We speculate that this result stems from TFMAN's training 
on $48\times 48$ patches, allowing the model to exhibit 
enhanced adaptability when applying non-local operations 
to blocks of the same size. Consequently, SRNL with 
a $48\times 48$ division size even outperforms the original non-local.
Additionally, the matching results of SRNL are visually 
presented in Figure \ref{fig_SRNL_visualization}.
While some blocks display subtle grid-like artifacts, 
these are easily mitigated through convolution operations 
and do not manifest in the final SR outcomes.

% Review 3
\begin{table}[!t]
    \renewcommand{\arraystretch}{0.6} %表示几倍行高
    \caption{
      Comparisons among SRNL and other lightweight non-locals. 
      PSNR results and total inference time on Set5 are reported after 
      5000 epochs of training. 
      The memory is the video memory required for the modules to 
      input a 128x128 feature map with 128 channels.
    } 
    \label{table_SRNL_comparison}
    \centering
    \tiny
  
    \begin{tabular}{|*{5}{@{}c@{}|}} %表格格式在此定义
      \hline 
      Module             &Non-Local\cite{Non_local} &NLSA\cite{SR:NLSN} &MDTA\cite{Restormer} & SRNL(Ours) \\
      \hline
      Inference Time (s) &1.05                      &0.91               &0.79                 &0.89\\ 
      Memory (GB)        &3.0                       &2.2                &1.1                  &1.9 \\
      PSNR (dB)	         &32.19                     &32.24              &32.09                &32.20 \\
      \hline          
      \end{tabular} 
  \end{table}
     
Finally, we conduct a comparison with other lightweight non-local 
approaches to showcase the strength of SRNL.
We use the `R2' structure in Table \ref{table_ablation_comparison} 
as the training carrier for x4 SR and set the model's 
recurrent depth to 4 to expedite training. 
As shown in Table \ref{table_SRNL_comparison},
NLSA \cite{SR:NLSN} comprehensively improves non-local in performance, 
inference speed and memory consumption.
Although SRNL's performance doesn't match that of NLSA, 
it has the same reasoning speed as NLSA and lower 
memory consumption.
Moreover, in contrast to NLSA that requires multiple complex hashes, 
SRNL only requires a straightforward division of the feature 
map during inference, which is easier to implement.

\section{Conclusion}
In this paper, we propose Trainable Feature 
Matching Attention Network (TFMAN) for SISR,
which incorporates the innovative Trainable Feature Matching (TFM) and 
Same-size-divided Region-level Non-Local (SRNL) modules.
Taking inspiration from early dictionary learning methods, 
TFM integrates an explicit feature learning mode into CNNs.
Trainable feature sets are included in TFM to explicitly learn the features 
through feature matching. 
This expansion of the CNN's representation mode enhances its 
representation capability and SR performance.
SRNL, a lightweight version of non-local operations, 
performs non-local computations independently in parallel 
on uniformly divided blocks of the input feature map.
SRNL significantly reduces computation and memory requirements 
compared to the original non-local, while achieving slightly 
improved super-resolution performance.
To enhance parameter utilization, we employ a recurrent 
convolutional network as the backbone of our proposed TFMAN.
We conduct comprehensive comparative experiments with BI, BD and DN
degradation models on benchmark datasets.
Our proposed TFMAN outperforms most of the state-of-the-art methods 
in terms of SR evaluation metrics and visual quality, 
while taking relatively fewer training parameters.
Furthermore, ablation study and module explorations are carried out
to verify the effectiveness of the proposed modules.

In general, TFM provides a new perspective for CNN-based image restoration, 
while SRNL provides a simple and effective practice for lightweight 
improvement of non-local. 
After simple modification, experts can adopt TFM 
or SRNL into other relevant vision tasks, such as image denoising, 
deblurring and defogging, etc., to reduce model resource 
consumption and improve performance.
Since the number of model parameters determines the upper bound of 
information carrying, it determines the upper bound of model performance. 
Although our model has achieved excellent SR performance with relatively 
few parameters, we believe there is still potential 
for further enhancement.
Therefore, in the future, we decide to do further research on model compression.

\section{Acknowledgements}
This work is supported by the National Natural Science Foundation of China,
(No. 61502220). In addtion, this work is supported by 
Science and Technology Commission of Shanghai 
Municipality (No. 19511105103).

% \appendix

\bibliographystyle{elsarticle-num-names} 
\bibliography{bibs/ElsevierAbrv, bibs/references}

\end{document}